\documentclass[lettersize,journal]{IEEEtran}
\usepackage{amsmath,amsfonts}
\usepackage{algorithm}
\usepackage{textcomp}
\usepackage{url}
\usepackage{verbatim}
\usepackage{graphicx}
\usepackage{cite}

\makeatletter%
\newcommand\hyperfootnotetext[1]{%
 {%
  \renewcommand{\@makefnmark}{}%
  \renewcommand{\leavevmode}{}%
  \renewcommand{\hyper@linkstart}[2]{}%
  \renewcommand{\hyper@linkend}{}%
  \footnote{#1}%
 }%
}%
\makeatother%
\usepackage{multirow}
\usepackage{booktabs}
\usepackage{verbatim}

\newcommand*{\Matrix}[1]{\ensuremath{\boldsymbol{#1}}}
\usepackage{hyperref}
\usepackage{tablefootnote}
\usepackage[noend]{algorithmic}

\begin{document}

\title{Unsupervised Feature Based Algorithms for Time Series Extrinsic Regression}

\author{David Guijo-Rubio
\thanks{}
\thanks{}}

\author{David Guijo-Rubio$^*$,
        Matthew Middlehurst,
        Guilherme Arcencio,
        Diego Furtado Silva,
        Anthony Bagnall
\IEEEcompsocitemizethanks{
\IEEEcompsocthanksitem $^*$: Corresponding author.%
\IEEEcompsocthanksitem D. Guijo-Rubio is with both the Department of Computer Science and Numerical Analysis, University of Cordoba, 14071 Cordoba, Spain and the School of Computing Sciences, University of East Anglia, Norwich, Norfolk, United Kingdom. E-mail: dguijo@uco.es%
\IEEEcompsocthanksitem M. Middlehurst and A. Bagnall are with the School of Computing Sciences, University of East Anglia, Norwich, Norfolk, United Kingdom. E-mail: \{m.middlehurst,ajb\}@uea.ac.uk
\IEEEcompsocthanksitem G. Arcencio is with the Department of Computing, Universidade Federal de São Carlos, São Carlos, Brazil. E-mail: garcencio@estudante.ufscar.br
\IEEEcompsocthanksitem D. Furtado Silva is with the Institute of Mathematics and Computer Sciences, Universidade de São Paulo, São Paulo, Brazil. E-mail: diegofsilva@usp.br}
\thanks{Manuscript received X; revised X.}}


\maketitle

\begin{abstract}
Time Series Extrinsic Regression (TSER) involves using a set of training time series to form a predictive model of a continuous response variable that is not directly related to the regressor series. The TSER archive for comparing algorithms was released in 2022 with 19 problems. We increase the size of this archive to 63 problems and reproduce the previous comparison of baseline algorithms. We then extend the comparison to include a wider range of standard regressors and the latest versions of TSER models used in the previous study. We show that none of the previously evaluated regressors can outperform a regression adaptation of a standard classifier, rotation forest. We introduce two new TSER algorithms developed from related work in time series classification. FreshPRINCE is a pipeline estimator consisting of a transform into a wide range of summary features followed by a rotation forest regressor. DrCIF is a tree ensemble that creates features from summary statistics over random intervals. Our study demonstrates that both algorithms, along with InceptionTime, exhibit significantly better performance compared to the other 18 regressors tested. More importantly, these two proposals (DrCIF and FreshPRINCE) models are the only ones that significantly outperform the standard rotation forest regressor.
\end{abstract}


\section{Introduction} \label{sec:introduction}
Time series analysis is a popular topic in machine learning and data mining research. Thousands of research papers in this field have been published in the last decade. Various algorithms have been proposed for disparate tasks across a wide range of applications. The main reason for this development is the increased ability to store data over time and the spread of cheap sensor technology to most fields of science. For example, solar panels depend on sensors to maximise their potential (e.g. to tilt the solar panel so that the sun shines directly on it) and hospitals routinely record and store patient data such as vital signs. This vast wealth of data offers great potential for data mining.

Two of the most researched time series machine learning/analysis tasks are classification \cite{bagnall2017great, middlehurst2023bake, ruiz2021great} and forecasting \cite{makridakis2008forecasting}. Time Series Classification (TSC) involves building a predictive model from (possibly multivariate) time series for a categorical target variable. TSC differs from standard classification in that the discriminatory features are often in the shape of the series or the autocorrelation. Forecasting consists of predicting (usually numeric) values based on past observations. Forecasting is usually approached through a model-based algorithm (e.g., autoregressive or exponential smoothing) or by reducing the forecasting problem to a regression problem through a sliding window then using deep learning or a global model such as XGBoost.

\begin{figure}[htb]
\centering
\includegraphics[width=\columnwidth, trim={0cm 0.2cm 0cm 0.2cm},clip]{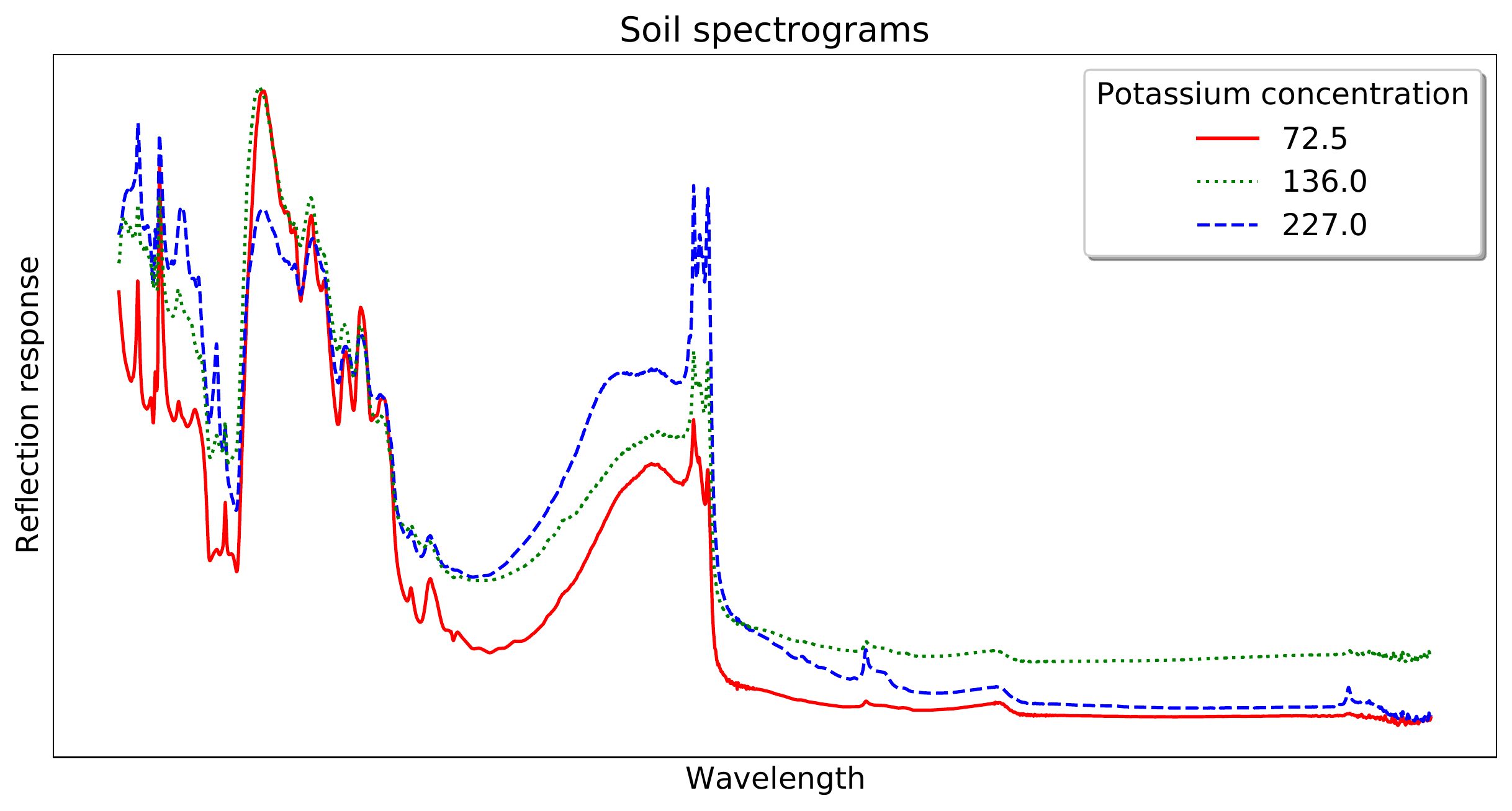}
\caption{Examples of soil spectrograms used to predict potassium concentration in the PotassiumConcentration dataset.}
\label{fig:datasetafricasoil}
\end{figure}

Tan \textit{et al.} in \cite{tan2021time} formally specified a related, but distinct, type of time series regression problem: Time Series {\bf Extrinsic} Regression (TSER). Rather than being derived from a forecasting problem, TSER involves a predictive model built on time series to predict a real-valued variable distinct from the training input series. For example, Figure \ref{fig:datasetafricasoil} shows soil spectrograms which can be used to estimate the potassium concentration. Ground truth is found through expensive lab based experiments that take some time. Spectrograms (ordered data series we treat as time series) are cheap to obtain and the data can be collected in any environment. An accurate regressor from spectrogram to concentration would make land and crop management more efficient. 

TSER is related to TSC as traditional regression is to classification: the only difference is that the target variable is real-valued rather than categorical. The distinction {\em extrinsic} is required because of the prevalence of the term time series regression in the forecasting literature to mean {\em reduce forecasting to regression through a sliding window}.

The first benchmarking work for TSER~\cite{tan2021time} introduced an archive of 19 TSER problems, including four univariate and 15 multivariate datasets. They performed an experimental comparison of the performance of 13 algorithms on these data. The two algorithms adapted from the TSC literature, the RandOm Convolutional KErnel Transform (ROCKET)~\cite{dempster2020rocket} and the deep learner InceptionTime~\cite{ismail2020inceptiontime}, were top-ranked. However, there was no significant difference in Root Mean Square Error (RMSE) between the ten best-performing algorithms, possibly because of the relatively small number of datasets and the conservative nature of the adjustment for multiple tests used. The abstract of~\cite{tan2021time} states that {\em ``we show that much research is needed in this field to improve the accuracy of ML models [for TSER]''}. 
Despite the paper's popularity and the identification of a clear need for novel research, there has been little or no progress in addressing this challenge. We have responded to this call to arms and developed and assessed a range of TSER algorithms. We have proposed new algorithms that are significantly better than the ones evaluated in~\cite{tan2021time}.

Our starting point to TSER is to adapt TSC algorithms for regression. The ROCKET family of classifiers all involve transformations using randomised convolutions and a pooling operation followed by a linear or ridge classifier. The original ROCKET was converted to TSER by switching the classifier for a ridge regressor. We extend this to consider a more recent ROCKET variant, MultiROCKET~\cite{tan2022multirocket}. Deep learning algorithms are also simple to adapt, and we extend the previous study to include a convolutionary neural network in addition to an ensemble regression version of InceptionTime~\cite{ismail2020inceptiontime}.

An alternative approach to TSC is to use a large number of unsupervised summary features as a transform. A review of a range of alternatives~\cite{middlehurst2022freshprince} found that the Fresh Pipeline with RotatIoN forest Classifier (FreshPRINCE) was the best transform pipeline for TSC. The FreshPRINCE uses the Time Series Feature Extraction based on Scalable Hypothesis Tests (TSFresh)~\cite{christ2018time} followed by a Rotation Forest (RotF) classifier~\cite{rodriguez2006rotation}. We implement the FreshPRINCE for TSER.

Interval-based classifiers also extract unsupervised features, but they do so by ensembling pipelines with randomly selected intervals and a fast base classifier. The first interval-based approach for TSC was the Time Series Forest (TSF) \cite{deng2013time}. TSF generates a set of random intervals and concatenates each interval's mean, standard deviation and slope to make a unique feature space for every base classifier. A version of TSF for regression is available in the aeon toolkit\footnote{\url{https://github.com/aeon-toolkit/aeon}.}. The Canonical Interval Forest (CIF)~\cite{middlehurst2020canonical}, and the subsequent Diverse Representation Canonical Interval Forest (DrCIF)~\cite{middlehurst2021hive}, adopt a similar model to the TSF but use different summary features and data representations. CIF uses the Canonical Time Series Characteristics (Catch22) \cite{lubba2019catch22} feature set. Details of these transformation-based algorithms and how we have adapted them to TSER are provided in Section~\ref{sec:unsup}. Our main contributions can be summarised as follows: 
\begin{enumerate}
    \item We provide 44 new datasets to the TSER archive, including 24 univariate and 20 multivariate datasets, to take the archive to 63 datasets;
    \item We repeat the study from~\cite{tan2021time} on these new data to examine whether the conclusions translate to the larger collection;
    \item We implement recently proposed convolutional-based, feature-based, interval-based, and deep learning-based TSC algorithms to TSER;
    \item We conduct an extensive experimental study using 21 regressors and demonstrate that feature-based and interval-based regressors, on average, achieve a significantly better RMSE than any other assessed algorithms;
    \item We carry out a comprehensive analysis and an ablative study of the two best proposed approaches: FreshPRINCE and DrCIF.
    \item We provide open source implementations of scikit-learn compatible implementations, clear guidance on reproducibility, and detailed results on the associated repository\footnote{\url{https://tsml-eval.readthedocs.io/en/latest/publications/2023/tser_archive_expansion/tser_archive_expansion.html}}.
\end{enumerate}

The rest of this paper is structured as follows. In Section \ref{sec:background}, the background and related works are exposed.
Section \ref{sec:unsup} describes our new TSER algorithms in detail. In Section \ref{sec:methods}, we give an overview of the new archive and describe our experimental setup. In Section \ref{sec:results}, experimental results for the 21 approaches applied to the total of 63 datasets are presented. Section \ref{sec:analysis} looks at these results in more detail. Finally, Section \ref{sec:conclusions} summarises our findings and highlights future work.

\section{Background and related work} \label{sec:background}


TSER aims to create a mapping function between a time series and a scalar value \cite{tan2021time}. A time series is composed of real-valued ordered observations. Formally, a univariate time series of length $m$ is defined as $\mathbf{s} = \{s_1, s_2, \ldots, s_m\}$. A multivariate time series with $d$ channels is specified as $\mathbf{S} = \{\mathbf{s}_1, \mathbf{s}_2, \ldots, \mathbf{s}_d\}$, where $\mathbf{s}_k = \{s_{1, k}, s_{2, k}, \ldots, s_{m, k}\}$. We assume we have a list $\mathbf{S}$ containing $n$ time series. Hence, $s_{i,j,k}$ represents the $j$-th observation of the $i$-th case for the $k$-th channel. A dataset $D$ is composed of $n$ time series samples and an associated response variable, $D = \{\mathbf{S}, \mathbf{y}\}$, where $\mathbf{y} = \{y_1, y_2, \ldots, y_n \}$ are the output continuous values, i. e. the input time series $\mathbf{s}_i$ is associated to the output $y_{i}$.

A TSER model is a mapping function $\mathcal{T} \rightarrow \mathcal{R}$, where $\mathcal{T}$ is the space of all time series and $\mathcal{R}$ a continuous value. A TSER model is trained on a dataset $D_{TRAIN}$ and evaluated on an independent test dataset $D_{TEST}$.

TSER shares some similarities with Scalar-on-Function Regression (SoFR)~\cite{goldsmith2014estimator}, a functional regression model where basis models are applied to the series prior to regression, i. e. its goal is to fit a regression model with scalar responses and functional data points as predictors~\cite{reiss2017methods}. ~\cite{tan2021time} used two SoFR models in their comparison based on~\cite{goldsmith2014estimator}. These were Functional Principal Component Regression (FPCR) and FPCR with B-splines (FPCR-Bs). 

Time series forecasting is often reduced to a regression window through the application of a sliding window to form time series $\mathbf{S}$ and a forecast horizon specifying how to select the target $\mathbf{y}$. The most common techniques used in Time Series Forecasting Regression (TSFR) include deep learning variants and global models, where channels are concatenated and a standard regressor such as Random Forest or XGBoost is applied.


Our primary source for TSER~\cite{tan2021time} compared three standard regression algorithms (Support Vector Regressors (SVR)~\cite{drucker1996support}, Random Forest (RandF)~\cite{breiman2001random}, and eXtreme Gradient Boosting (XGBoost)~\cite{chen2016xgboost}); two $k$-Nearest Neighbours models using Euclidean and Dynamic Time Warping distances using both one and five neighbours; three deep learning approaches (Fully Convolutional Neural Network (FCN) \cite{wang2017time}, Residual Network (ResNet)~\cite{he2016deep} and InceptionTime ~\cite{ismail2020inceptiontime}); two functional analysis approaches (FPCR and FPCR with B-splines~\cite{goldsmith2014estimator}); and ROCKET \cite{dempster2020rocket}. The standard regressors adopt the approach of global forecasting regressors: time series are flattened into a vector concatenating all the channels. Hence, a multivariate time series of length $m$ and $d$ channels is converted into a single vector of length $m \times d$. Subsequently, there have been very few algorithmic advances for TSER. Most novel developments are domain specific and not aimed at TSER as a whole. Among them, a Linear Space State Layers (LSSL) model~\cite{gu2021state} has been tested on three TSER datasets, achieving low error metrics. Another state-space model, Liquid-S4~\cite{hasani2022liquid} has also been evaluated on those three datasets, and claims better results. An architecture based on Graph Neural Networks called TISER-GCN~\cite{bloemheuvel2022graph} has been applied to seismic data as an extrinsic regression task. In a similar context, \cite{siddiquee2022septor} introduces Septor, a hierarchical neural network model developed to estimate the depth of seismic events from waveform data, i.e. a domain-specific extrinsic regression task. ROCKET-XGBoost~\cite{bayani2022essay} has been the only novel algorithm evaluated on the 19 TSER archive datasets, but it offered no significant improvement over the algorithms evaluated in~\cite{tan2021time}.

\subsection{Time Series Classification (TSC) algorithms}
There are a plethora of algorithms for TSC that have been compared in reproducible comparative studies~\cite{bagnall2017great,middlehurst2023bake,ruiz2021great}. Broadly, algorithms can be grouped into how they represent time series. We provide a very brief overview with a focus on how classifiers have been or could be adapted to TSER.

\textbf{Distance-based} classifiers use a distance function in conjunction with an algorithm such as Nearest Neighbour (NN) classifier. The two most commonly used distance functions are Euclidean distance and Dynamic Time Warping (DTW). A NN classifier can trivially be adapted for regression by averaging over the target variable of the NN. For multivariate data, using terminology presented in ~\cite{shokoohi17generalizing}, DTW can either be independent (find DTW distance on each channel separately them sum the values) or dependent (use all channels in the point wise distance calculation).

\textbf{Feature-based} algorithms transform series into features using unsupervised descriptive statistics, then complete the pipeline with a classifier trained on the new feature set.

\textbf{Interval-based} classifiers are an extension of feature-based pipeline classifiers where rather than form summary features over the whole series, they concatenate features found over different intervals. They then form an ensemble over different randomised intervals rather than use a single estimator. Together, we group feature-based and interval-based approaches together as unsupervised feature-based classifiers. Adapting these algorithms to TSER is our primary research goal, so we cover this topic in more detail in Section~\ref{sec:unsup}.

\textbf{Kernel/convolution-based} models find convolutions from the space of all possible subseries and use them to create features through a form of pooling operation. The most popular approach, ROCKET~\cite{dempster2020rocket} generates random convolutions and is used in conjunction with a ridge classifier in a pipeline. It was adapted for TSER by simply changing the ridge classifier for a ridge regressor~\cite{tan2021time}. More recently, MultiROCKET~\cite{tan22multirocket} was proposed as an improved version of ROCKET. ROCKET uses two pooling operations to generate features: max pooling and the percentage of positive values. MultiROCKET adds three new pooling operations: mean of positive values, mean of indices of positive values, and longest stretch of positive values. It also extracts features from first order differences in addition to the raw data. We adapt MultiROCKET for TSER in exactly the same way as ROCKET.

{\bf Deep learning} continues to be popular for TSC~\cite{ismail2019deep}, although to our knowledge InceptionTime~\cite{ismail2020inceptiontime} is still the best performing deep learner. The study in~\cite{tan2021time} used Residual Networks (ResNet), Fully Connected Networks (FCN), and InceptionTime. The original InceptionTime paper~\cite{ismail2020inceptiontime} proposed an ensemble of five InceptionTime classifiers to obtain the final results. However,~\cite{tan2021time} used a single InceptionTime model for TSER. We evaluate both a single InceptionTime (Inception) and an InceptionTime Ensemble (InceptionTimeE) faithful to the TSC version for TSER. We also evaluate the Convolutional Neural Network (CNN) regressor based on the classifier described in~\cite{zhao2017cnn}.

\textbf{Shapelet-based} approaches~\cite{bostrom17binary, ye11shapelets} base classification on the presence of selected phase-independent subseries found from the training data. For classification, shapelets are assessed with a supervised measure such as information gain. Furthermore, the most accurate shapelet-based approaches~\cite{middlehurst2021hive} evaluate shapelets with a one vs many approach and balance the search procedure between classes to improve diversity. Adapting shapelets for TSER requires significant internal changes and design decisions, being beyond the scope of this paper.

\textbf{Dictionary based} algorithms use a bag of words-like approach to base classification on the number of occurrences of approximated subseries (patterns). The most successful dictionary-based classifiers~\cite{schafer15boss, schafer2023weasel, middlehurst20temporal} involve a degree of supervised selection using accuracy for filtering/weighting or feature selection and their adaptation for TSER is also beyond the scope of this paper.

\section{Unsupervised Feature-Based Regressors}
\label{sec:unsup}

Approaches which extract features from time series in an unsupervised process have been shown to perform well in classification scenarios. ROCKET~\cite{dempster2020rocket} and CIF~\cite{middlehurst2020canonical} perform as well as or better than single representation approaches such as shapelet or dictionary algorithms. These algorithms also have the benefit of lower complexity, essentially consisting of transform to classifier pipelines or an ensemble of pipelines. ROCKET and CIF~\cite{middlehurst2020canonical} were also top ranked for Multivariate TSC (MTSC) in a recent survey~\cite{ruiz2021great}.

We describe the features extracted and our adaptations for two additional algorithms based on unsupervised transformations. The first is the FreshPRINCE~\cite{middlehurst2022freshprince}, a pipeline using the TSFresh~\cite{christ2018time} feature set. The second is DrCIF~\cite{middlehurst2021hive}, an interval-based ensemble.

\subsection{FreshPRINCE} 
\label{sec:fp}

FreshPRINCE is a pipeline algorithm for regression with two components: the TSFresh feature extraction algorithm that transforms the input time series into a feature vector, and then a Rotation Forest (RotF)~\cite{rodriguez2006rotation} estimator that builds a model and makes label predictions. TSFresh~\cite{christ18time} is a collection of just under 800 features that can be extracted from time series data. While the features can be used on their own, a feature selection method called Fresh is provided to remove irrelevant features. FreshPRINCE does not make use of this feature extraction, however, keeping the transformation process unsupervised and allowing the RotF decide the worth of features. TSFresh is generally popular within the data science community, and has shown to perform better than other unsupervised transformation pipelines on classification problems as part of FreshPRINCE~\cite{middlehurst2022freshprince}.


RotF is an ensemble of tree classifiers which has been shown to accurately make predictions for problems where the attributes are continuous~\cite{bagnall2018rotation}. The classifier has been used as a benchmark and as a part of other pipeline classifiers in TSC~\cite{bagnall2017great,middlehurst2021hive}, and performed better than a ridge classifier and XGBoost~\cite{chen2016xgboost} when paired with unsupervised transforms for TSC~\cite{middlehurst2022freshprince}. Full descriptions of the RotF algorithm are available in~\cite{rodriguez2006rotation} and~\cite{bagnall2018rotation}. RotF is easily adaptable for regression: the implementation we developed removes class subsampling~\cite{pardo2013rotation}, replaces the C4.5 decision tree with a Classification and Regression Tree (CART)~\cite{breiman1984classification}, and averages the label predictions for each tree in the forest. The full TSFresh transformation and altered RotF make up our FreshPRINCE adaptation for TSER.

\subsection{DrCIF}
\label{sec:drcif}

Interval-based techniques select phase-dependent intervals of fixed offsets from which to extract summary features. These intervals share their position for all time series, with the aim of discovering discriminatory features from particular locations in time. Most interval techniques take the form of a forest of decision trees, using different intervals to achieve diversity in the ensemble. While some interval forests do make use of supervised feature extraction~\cite{cabello2020fast}, TSF~\cite{deng2013time} and DrCIF~\cite{middlehurst2021hive} are completely unsupervised in their method for selecting intervals and extracting features from said intervals. All that we change for TSER from the classification implementation is a swapping of the tree algorithm used. TSF can be adapted for the regression task in the same way.

\begin{figure*}[htpb]
        \centering
        \begin{tabular}{cc}
            \includegraphics[width=0.61\textwidth]{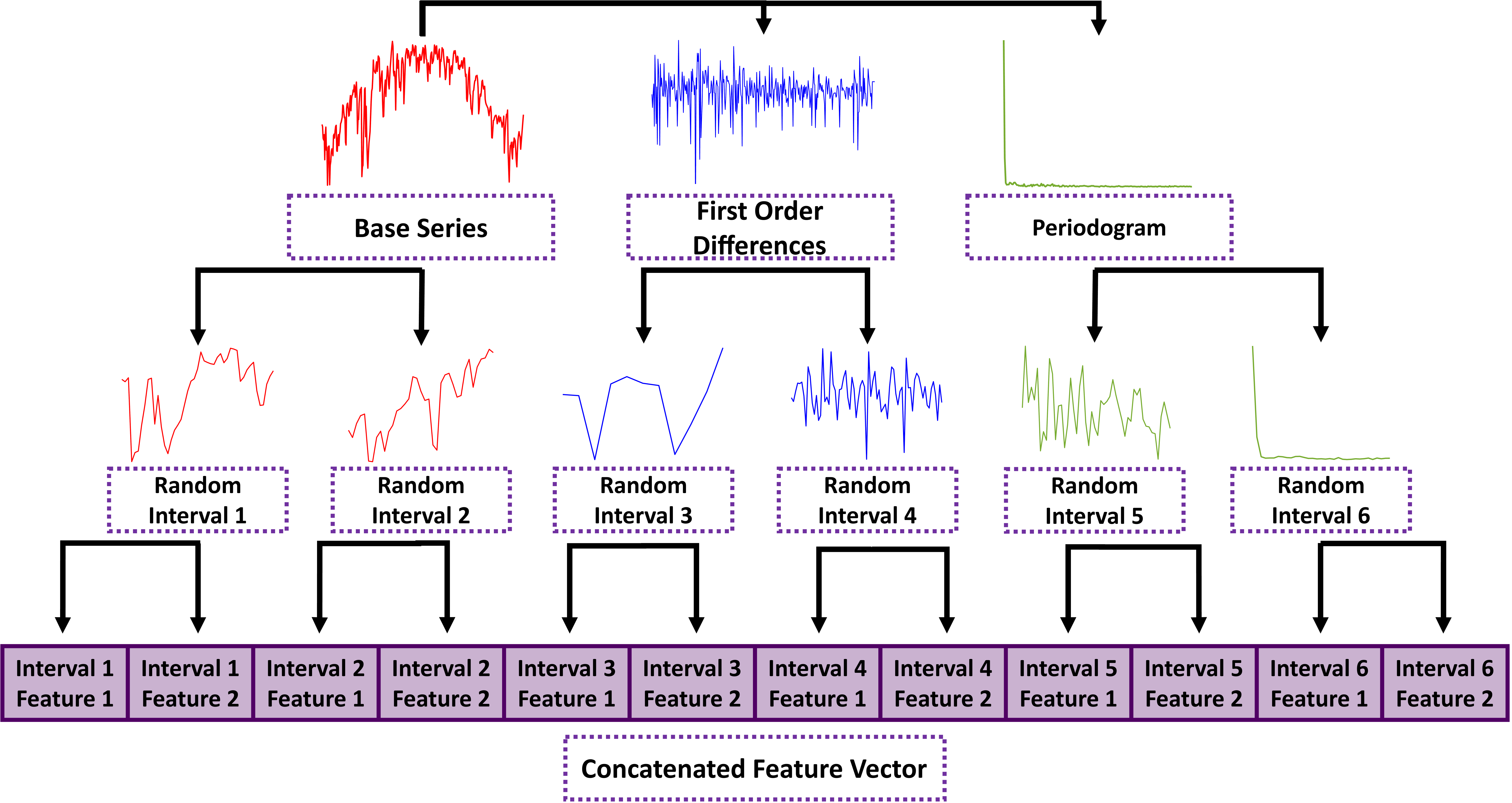} & 
            \includegraphics[width=0.35\textwidth]{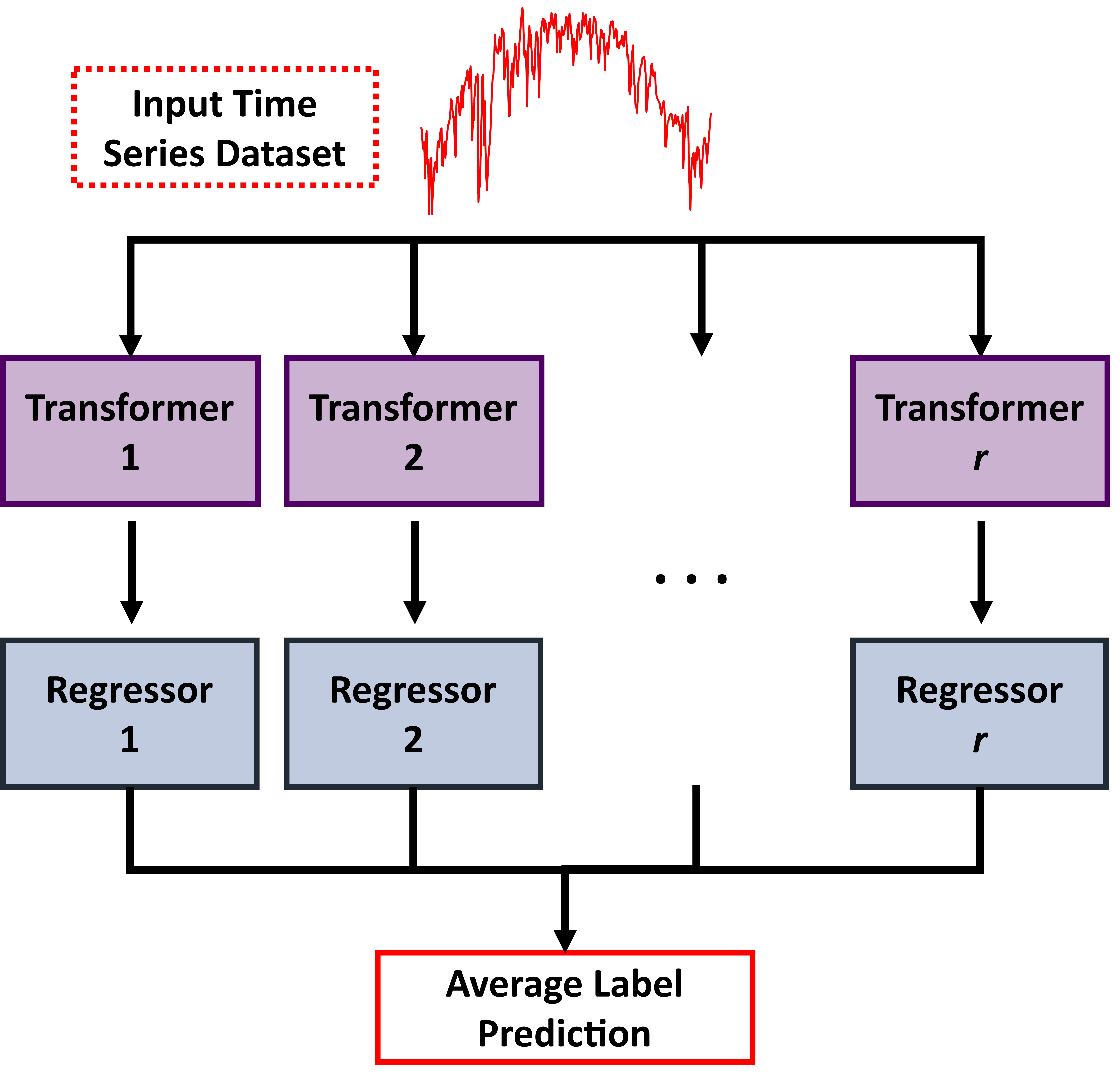} \\
        \end{tabular}
        \caption{Diagrams visualising the DrCIF transformation (left) and DrCIF ensemble structure (right).}
        \label{fig:DrCIF}
    \end{figure*}

From a series of length $m$, there are $m(m-1)/2$ possible intervals when considering all interval lengths and positions. Even at small series lengths, it is unfeasible to extract features from or evaluate all possible intervals. To solve the issue of which intervals from this pool to select, DrCIF uses a random forest based approach. An ensemble of CART regressors is formed, built on the output of different random interval transformations. Algorithm~\ref{alg:DrCIF} describes the full build process for DrCIF. The transformation has three steps. First, the base time series is split into three series representations: the original time series, the first order differences of the series, and the periodogram of the series (characterised in line 3 of Algorithm~\ref{alg:DrCIF}). The differences and periodogram series-to-series transformations have shown to provide useful information in classification approaches~\cite{flynn2019contract,cabello2020fast,tan2022multirocket,keogh2001derivative}. Then, a different transform is created for each base regressor. First, a pool of $a$ features is selected from a candidate pool of 29 features (line 6). DrCIF makes use of the CAnonical Time series CHaracteristics (Catch22)~\cite{lubba2019catch22}. Catch22 is a diverse set of 22 features filtered from the 7000+ available in the Highly Comparative Time Series Analysis (HCTSA) toolbox~\cite{fulcher17hctsa}. The Catch22 features were selected for use on normalised data, but we do not make that assumption. Hence, seven additional summary statistics are also candidates: the mean, standard-deviation, slope, median, interquartile range, min, and max. Then, for each data representation, a set of $k$ random intervals are selected (lines 10-13), and the $a$ unsupervised features are calculated and concatenated from a randomly selected channel (lines 13-15). Finally, a CART tree is trained on the feature set unique to each ensemble member. Figure~\ref{fig:DrCIF} visualises the transformation (left) and ensemble (right) process for DrCIF. Predictions for new cases are found by averaging the predictions of the base regressors.
    \begin{algorithm}[htb]
	\caption{DrCIF(A list of $n$ cases of length $m$ with $d$ channels, $\Matrix{T}=(\Matrix{X},\Vec{y})$)}
    \label{alg:DrCIF}
        \begin{algorithmic}[1]
            \REQUIRE the number of trees: $r$; the number of intervals per representation for each tree: $k$; and the number of attributes subsampled per tree: $a$ (default $r=500$, $k=4+(\sqrt{d}\sqrt{rm}$)/3, and $a=10$)  \COMMENT{ {\em where rm is the length of a representations series}}
            \STATE Let $\Vec{DT}$ = $(DT_1 \ldots DT_r)$ be the trees in the forest
            \STATE Let $\Matrix{V}$ be a $3 \times n \times  d$ matrix of series with variable length,  containing the base series, the periodograms and first order differences
            \STATE $\Matrix{V} \leftarrow transform(\Matrix{X})$ 
        	\FOR {$i \leftarrow 1$ to $r$}
                \STATE $\Matrix{X'} \leftarrow []$
                \STATE $\Vec{u} \leftarrow$ select $a$ random attribute indices $(u_1, \ldots , u_a)$
                \FOR {$re \leftarrow 1$ to $3$}
                    \STATE $\Matrix{F} \leftarrow$ initialize matrix of dimensionality $n \times (ak)$
            		\FOR {$j \leftarrow 1$ to $k$}
            		    \STATE $b = rand(1,|\Matrix{V}_{re}|-3)$ \COMMENT{ {\em interval position} }
            		    \STATE $l = rand(3,|\Matrix{V}_{re}|/2)$ \COMMENT{ {\em interval length} }
            		    \STATE $o = rand(1,d)$ \COMMENT{ {\em interval channel} }
            		    \FOR {$t \leftarrow 1$ to $n$}
            		        \FOR {$c \leftarrow 1$ to $a$}
            		            \STATE $\Matrix{F}_{t,a(j-1)+c} \leftarrow summaryStat(u_c,\Matrix{V}_{re,t,o},b,l)$
            		        \ENDFOR
            		    \ENDFOR
            		\ENDFOR
                    \STATE $\Matrix{X'} \leftarrow \Matrix{X'} + \Matrix{F}$ \COMMENT{ {\em concatenate feature vectors } }
        		\ENDFOR
        		\STATE $DT_i.buildCART(\Matrix{X'},\Vec{y})$
            \ENDFOR
        \end{algorithmic}
    \end{algorithm}

\section{Methodology} \label{sec:methods}

We summarise the new problems we have added, the regressors used in experiments, and a description of our experimental method. 

The current version of the TSER archive includes 19 different datasets. We have increased the number of datasets in the archive to 63. There are now 28 univariate problems and 35 multivariate, with number of channels ranging from 2 to 24. Dataset size range from 93 to over 16,000. 70\% are used for training, 30\% for testing. Series length ranges from 14 to 7500. Nine of the problems contain missing values and two have unequal-length series.

The new datasets have been taken from Kaggle competitions and other archives and repositories/websites associated with applied research. Table~\ref{tab:datasets_brief} summarises the gathered data. More details on the datasets are available on the associated repository. None of the datasets have been normalised. One of the new problems has unequal-length series. For experiments, keeping with the practice in~\cite{tan2021time}, missing values in the series are linearly interpolated, and unequal-length series are truncated to the minimum length series. Full descriptions, and both unequal and missing values series are available on the associated website.


\begin{table*}[htpb]
    \caption{New TSER datasets}
    \label{tab:datasets_brief}
    \resizebox{0.91\textwidth}{!}{
    \begin{tabular}{l|l}
    \toprule \toprule
    Name & Prediction problem (response variable) \\ \midrule
    \multicolumn{2}{c}{\textbf{Economic Analysis}}\\  \midrule
    
        \textbf{DailyOilGasPrices} & Daily gas price with time series of oil  prices\textsuperscript{\ref{dailyOilGasPrices}}.\\ \midrule

    \multicolumn{2}{c}{\textbf{Energy Monitoring}}\\  \midrule

        \textbf{Energy building predictors} & Estimate the energy consumption of different sorts on buildings\textsuperscript{\ref{EnergyBuilding}}.\\
    
        \textbf{OccupancyDetectionLight} & The average hourly occupancy of an office from sensor measurements \cite{datasets_CANDANEDO201628}.\\
        
        \textbf{SolarRadiationAndalusia} & The average hourly solar radiation from atmospherical measurements\textsuperscript{\ref{SolarRadiationAndalusia}}.\\

        \textbf{TetuanEnergyConsumption} & The daily average power consumption in three areas of Tetouan from atmospherical measurements\cite{salam2018comparison}.\\ 
        
        \textbf{WindTurbinePower} & The daily power output of a wind turbine based on time series of torque measurements\textsuperscript{\ref{WindTurbinePower}}.\\ \midrule

    \multicolumn{2}{c}{\textbf{Environment monitoring}}.\\  \midrule

        \textbf{AcousticContaminationMadrid} & The 1st percentile of sound pressure levels from LAeq\textsuperscript{\ref{AcousticContaminationMadrid}}.\\

        \textbf{Africa Soil Chemistry} & A set of 12 problems derived from the Africa Soil Information Service (AfSIS) Soil Chemistry
        \textsuperscript{\ref{AfricaSoilChemistry}}.\\ 

        \textbf{BeijingIntAirportPM25Quality} & The daily average of particulate matter in the Airport of Beijing from atmospherical data. \cite{liang2015assessing}.\\

        \textbf{DailyTemperatureLatitude} & The latitude of a city based on the annual time series of daily temperature\textsuperscript{\ref{DailyTemperatureLatitude}}.\\

        \textbf{DhakaHourlyAirQuality} & The Air Quality Index in Dhaka using localised particular matter time series\textsuperscript{\ref{DhakaHourlyAirQuality}}.\\

        \textbf{MadridPM10Quality} & The weekly average of particulate matter in the city of Madrid, Spain, from measurements of gases\textsuperscript{\ref{MadridPM10Quality}}.\\

        \textbf{MetroInterstateTrafficVolume} & The daily average traffic volume of a road in the USA from atmospherical variables\textsuperscript{\ref{MetroInterstateTrafficVolume}}.\\

        \textbf{ParkingBirmingham} & The daily occupancy rate from the hourly total number of parked cars \cite{stolfi2017predicting}.\\

        \textbf{PrecipitationAndalusia} & The yearly average of rainfall on Andalusia, Spain, from meterological measurements\textsuperscript{\ref{PrecipitationAndalusia}}.\\

        \textbf{SierraNevadaMountainsSnow} & The amount of snow based on temperature time series~\cite{datasets_SIERRANEVADA}.\\
        \midrule
    \multicolumn{2}{c}{\textbf{Equipment monitoring}}\\  \midrule

        \textbf{ElectricMotorTemperature} & The temperature of an electric motor based on time series of torque readings\textsuperscript{\ref{ElectricMotorTemperature}}.\\

        \textbf{LPGas/Methane MonitoringHomeActivity} & The liquefied petroleum and methane concentration from gas sensors~\cite{huerta2016online}.\\
        
        \textbf{GasSensorArray Ethanol/Acetone} & The concentrations of two analytes, acetone based on 16 metal-oxide sensors~\cite{datasets_ZIYATDINOV2015538}.\\
        
        
        \textbf{WaveTensionData} & The tension of a string based on wave elevation time series\textsuperscript{\ref{WaveTensionData}}.\\         \midrule

        \multicolumn{2}{c}{\textbf{Health Monitoring}}\\  \midrule

        \textbf{BarCrawl6min} & The transdermal alcohol content by using an accelerometer~\cite{killian2019learning}.\\

        \textbf{Covid19Andalusia} & The rate of deceased/contagions people from number of contagions in Andalusia, Spain~\cite{diaz2022covid}.\\ 

        \textbf{VentilatorPressure} & The pressure of the inspiratory solenoid valve from control input and output of the same valve\textsuperscript{\ref{VentilatorPressure}}.\\ 
        
        \midrule

        \multicolumn{2}{c}{\textbf{Sentiment Analysis}}\\  \midrule
         
        \textbf{Crypto Sentiment} & The sentiment of four cryptocurrencies based on the same days hourly price\textsuperscript{\ref{CryptoSentiment}}.\\
        
        \textbf{NaturalGasPriceSentiment} & Sentiment scores about natural gas prices~\cite{datasets_gas_sentiment} based on the daily natural gas prices.\\

\bottomrule \bottomrule

    \end{tabular}
    }
    
\end{table*}

\hyperfootnotetext{\label{dailyOilGasPrices}\textsuperscript{\ref*{dailyOilGasPrices}}\url{https://perma.cc/DAP4-JC2A}}

\hyperfootnotetext{\label{EnergyBuilding}\textsuperscript{\ref*{EnergyBuilding}}\url{https://perma.cc/PA63-7GVU}}


\hyperfootnotetext{\label{SolarRadiationAndalusia}\textsuperscript{\ref*{SolarRadiationAndalusia}}\url{https://perma.cc/CH23-UVRJ}}


\hyperfootnotetext{\label{WindTurbinePower}\textsuperscript{\ref*{WindTurbinePower}}\url{https://perma.cc/8X2R-PRUD}}

\hyperfootnotetext{\label{AcousticContaminationMadrid}\textsuperscript{\ref*{AcousticContaminationMadrid}}\url{https://perma.cc/9V27-BF3F}}

\hyperfootnotetext{\label{AfricaSoilChemistry}\textsuperscript{\ref*{AfricaSoilChemistry}}\url{https://perma.cc/TP5Y-KS6M}}


\hyperfootnotetext{\label{DailyTemperatureLatitude}\textsuperscript{\ref*{DailyTemperatureLatitude}}\url{https://perma.cc/3KPY-YHW2}}

\hyperfootnotetext{\label{DhakaHourlyAirQuality}\textsuperscript{\ref*{DhakaHourlyAirQuality}}\url{https://perma.cc/7865-ZAAD}}

\hyperfootnotetext{\label{MadridPM10Quality}\textsuperscript{\ref*{MadridPM10Quality}}\url{https://perma.cc/ZUS5-E26E}}

\hyperfootnotetext{\label{MetroInterstateTrafficVolume}\textsuperscript{\ref*{MetroInterstateTrafficVolume}}\url{https://perma.cc/B6FV-SLCG}}


\hyperfootnotetext{\label{PrecipitationAndalusia}\textsuperscript{\ref*{PrecipitationAndalusia}}\url{https://perma.cc/3APP-2L43}}


\hyperfootnotetext{\label{ElectricMotorTemperature}\textsuperscript{\ref*{ElectricMotorTemperature}}\url{https://perma.cc/A7FG-KFLT}}



\hyperfootnotetext{\label{WaveTensionData}\textsuperscript{\ref*{WaveTensionData}}\url{https://perma.cc/BAG4-W8SL}}




\hyperfootnotetext{\label{VentilatorPressure}\textsuperscript{\ref*{VentilatorPressure}}\url{https://perma.cc/RQW7-QH7L}}

\hyperfootnotetext{\label{CryptoSentiment}\textsuperscript{\ref*{CryptoSentiment}}\url{https://perma.cc/J6LK-99Q5}}


\subsection{Regression Algorithms}
The full list of the 21 regressors (with associated abbreviation) evaluated in Section~\ref{sec:results} is as follows:
\begin{itemize}
    \item Standard ML regressors:
    \begin{itemize}
        \item Ridge Regression (Ridge).
        \item Grid-search Support Vector Regression (Grid-SVR) \cite{drucker1996support}.
        \item Random Forest (RandF) \cite{breiman2001random}.
        \item Rotation Forest (RotF) \cite{rodriguez2006rotation}.
        \item eXtreme Gradient Boosting (XGBoost) \cite{chen2016xgboost}.
    \end{itemize}
    \item Functional Linear Models:
    \begin{itemize}
        \item Functional Principal Component Analysis (FPCR) \cite{goldsmith2014estimator}.
        \item FPCR with B-splines (FPCR-Bs)  \cite{goldsmith2014estimator}.
    \end{itemize}
    \item Convolutional-based:
    \begin{itemize}
        \item ROCKET \cite{dempster2020rocket}.
        \item MultiROCKET \cite{tan2022multirocket}.
    \end{itemize}
    \item Distance-based:
    \begin{itemize}
        \item $k$ Nearest neighbours with Euclidean distance ($k$NN-ed), with $k =\{1, 5\}$.
        \item $k$ Nearest neighbours with dynamic time warping distance ($k$NN-dtw), with $k =\{1, 5\}$.
    \end{itemize}
    \item Feature-based:
    \begin{itemize}
        \item FreshPRINCE \cite{middlehurst2022freshprince}.
    \end{itemize}
    \item Interval-based:
    \begin{itemize}
       \item Time Series Forest (TSF) \cite{deng2013time}.
         \item DrCIF~\cite{middlehurst2021hive}.
    \end{itemize}
    \item Deep Learning approaches:
    \begin{itemize}
        \item Convolutional Neural Network (CNN) \cite{ismail2019deep}.
        \item Fully Connected Neural Network (FCN) \cite{ismail2019deep}.
        \item Residual Neural Network (ResNet) \cite{ismail2019deep}.
        \item Single Inception Neural Network (Inception) \cite{ismail2020inceptiontime}.
        \item Inception Neural Network Ensemble (InceptionE) \cite{ismail2020inceptiontime}.
    \end{itemize}
\end{itemize}
Parameter settings for all algorithms are in the supplementary material. 

\subsection{Experimental Design}
Each dataset is provided with a default train/test split. We repeat every experiment 30 times to obtain a more significant analysis. The first experiment is with the default data. Subsequent experiments are conducted with data resampled by pooling the train and test and randomly partitioning the data with the same train/test proportions as the original. Performance is measured with the RMSE to conform with~\cite{tan2021time}. 
To compare regressors, we first average RMSE over all resamples. We use ranks in all statistical tests. 
For multiple regressors over multiple datasets we use an adaptation of the critical difference diagram~\cite{demsar06comparisons}, replacing the post-hoc Nemenyi test with a comparison of all classifiers using pairwise Wilcoxon signed-rank tests, and cliques formed using the Holm correction~\cite{garcia08pairwise,benavoli16pairwise}.
\section{Results}
\label{sec:results}
\label{sec:recreate}
\begin{figure*}[t]
    \centering
    \begin{tabular}{cc}
    \includegraphics[width=0.5\textwidth,trim={0cm -1cm 0cm 0cm},clip]{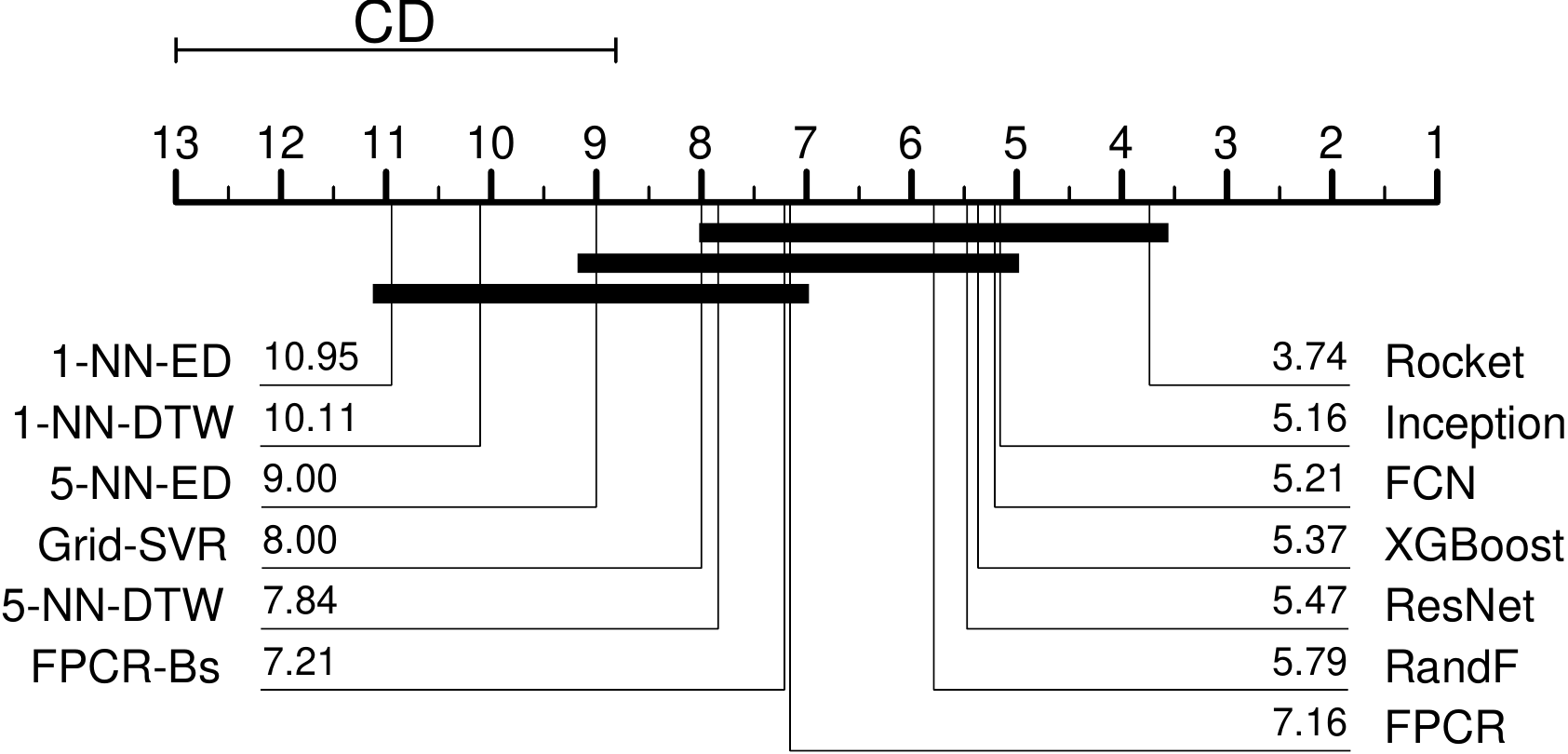} & 
    \includegraphics[width=0.5\textwidth,trim={2.5cm 2cm 0cm 0cm},clip]{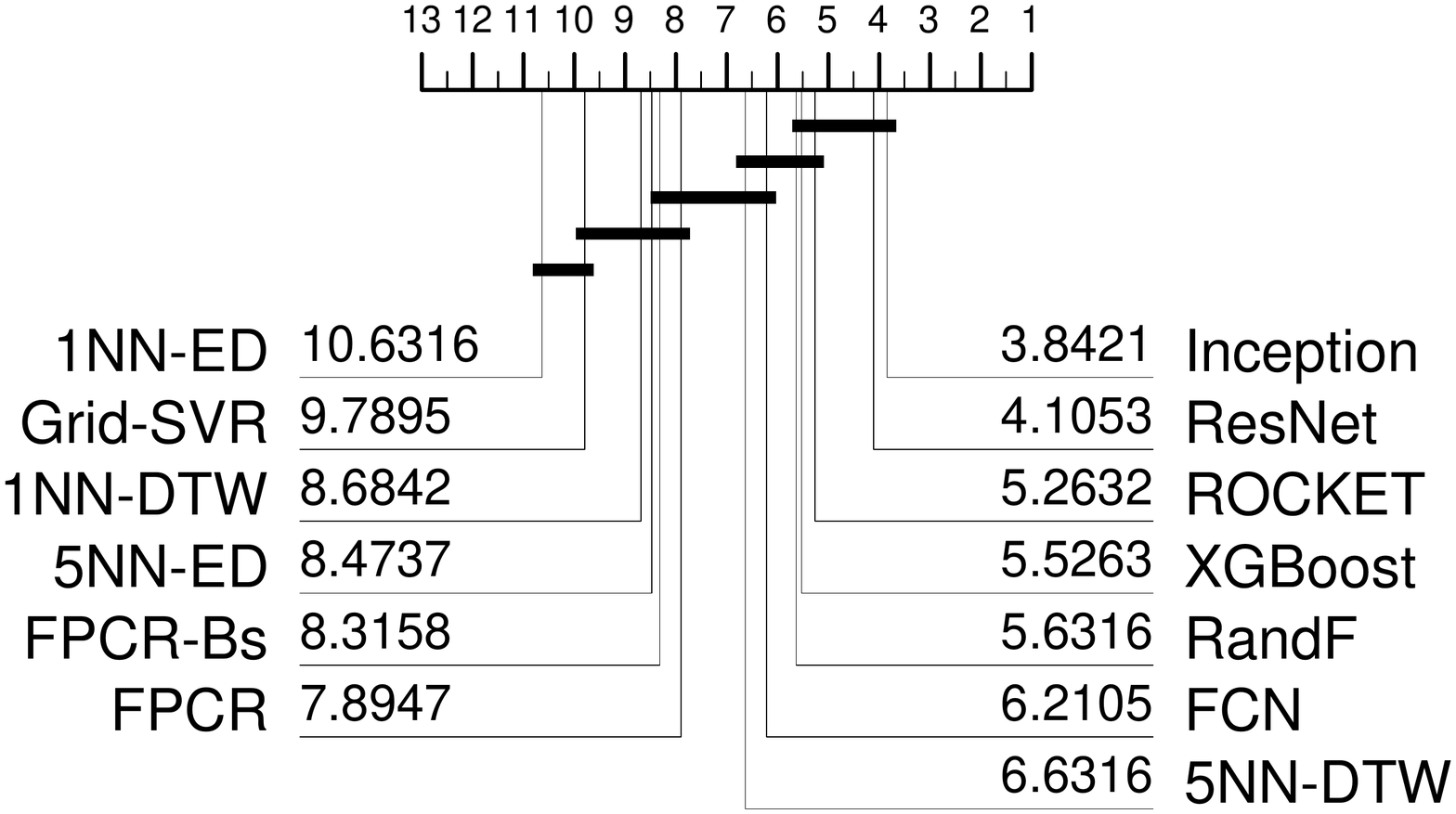} \\
    \end{tabular}
    \caption{Reproduction of the RMSE ranks on the original archive (19 datasets and 5 resamples). Left is the original image from \cite{tan2021time}. Right is our recreation.} 
    \label{fig:recreate1}
\end{figure*}
Our experiments are structured as follows. In Section~\ref{sec:recreate} we recreate the results presented in~\cite{tan2021time} on the original 19 datasets. We then extend the analysis to our larger collection of datasets to test whether the conclusions reached in~\cite{tan2021time} generalise to the new archive of 63 problems. In Section~\ref{sec:bench} we include two improvements for regressors used in the previous study and other regressors available in open source toolkits. In Section~\ref{sec:new} we compare the best performing regressors from the previous experiments to the new algorithms we are proposing, FreshPRINCE and DrCIF. 

\subsection{Recreating Results on the 19 TSER datasets}
\label{sec:recreate}
We ran the 13 regressors reported in~\cite{tan2021time} on the current 19 datasets in the archive, using five resamples, keeping in practise with the original work. Figure~\ref{fig:recreate1} shows a critical difference diagram of our results alongside the results presented in~\cite{tan2021time}.
Broadly, the ordering of algorithms is the same and the cliques are similar. There are some differences in the ordering, with ROCKET and FCN lower ranked and Inception and ResNet higher in our experiments than the original. We also have more diverse cliques. This is because our adjustment for multiple testing is less conservative than the one used in~\cite{tan2021time}, where a full Bonferonni adjustment is used rather than a Holm correction.
\begin{figure}[htb]
    \centering
    \includegraphics[width=0.475\textwidth,trim={3cm 2cm 2cm 5cm},clip]{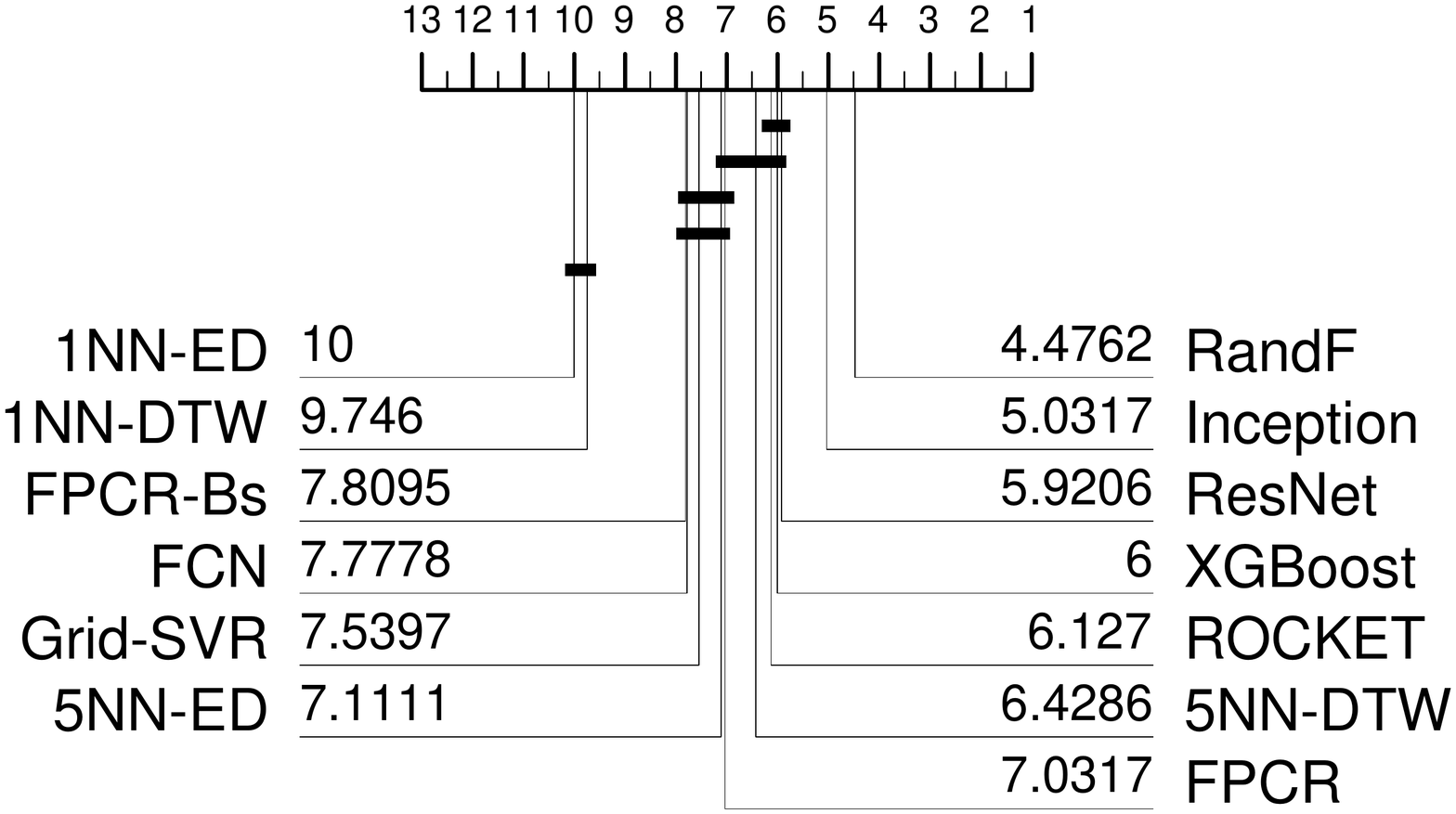}
    \caption{RMSE ranks for 13 Regressors used in~\cite{tan2021time} on 63 TSER datasets.} 
    \label{fig:recreate2}
\end{figure}

In Figure~\ref{fig:recreate2} we compare the 13 regressors used for Figure~\ref{fig:recreate1} on the larger archive of 63 datasets. From now on, we also extended the number of resamples from five to 30. We note that RandF is now the best, improving significantly on ROCKET, Inception, and ResNet. Hence, the time series specific methods previously proposed for TSER are not better than using an off the shelf regressor with concatenated features.

\subsection{Benchmarking the new TSER archive}
\label{sec:bench}
\begin{figure}[htb]
    \centering
    \includegraphics[width=0.475\textwidth,trim={3cm 3.3cm 0.4cm 5cm},clip]{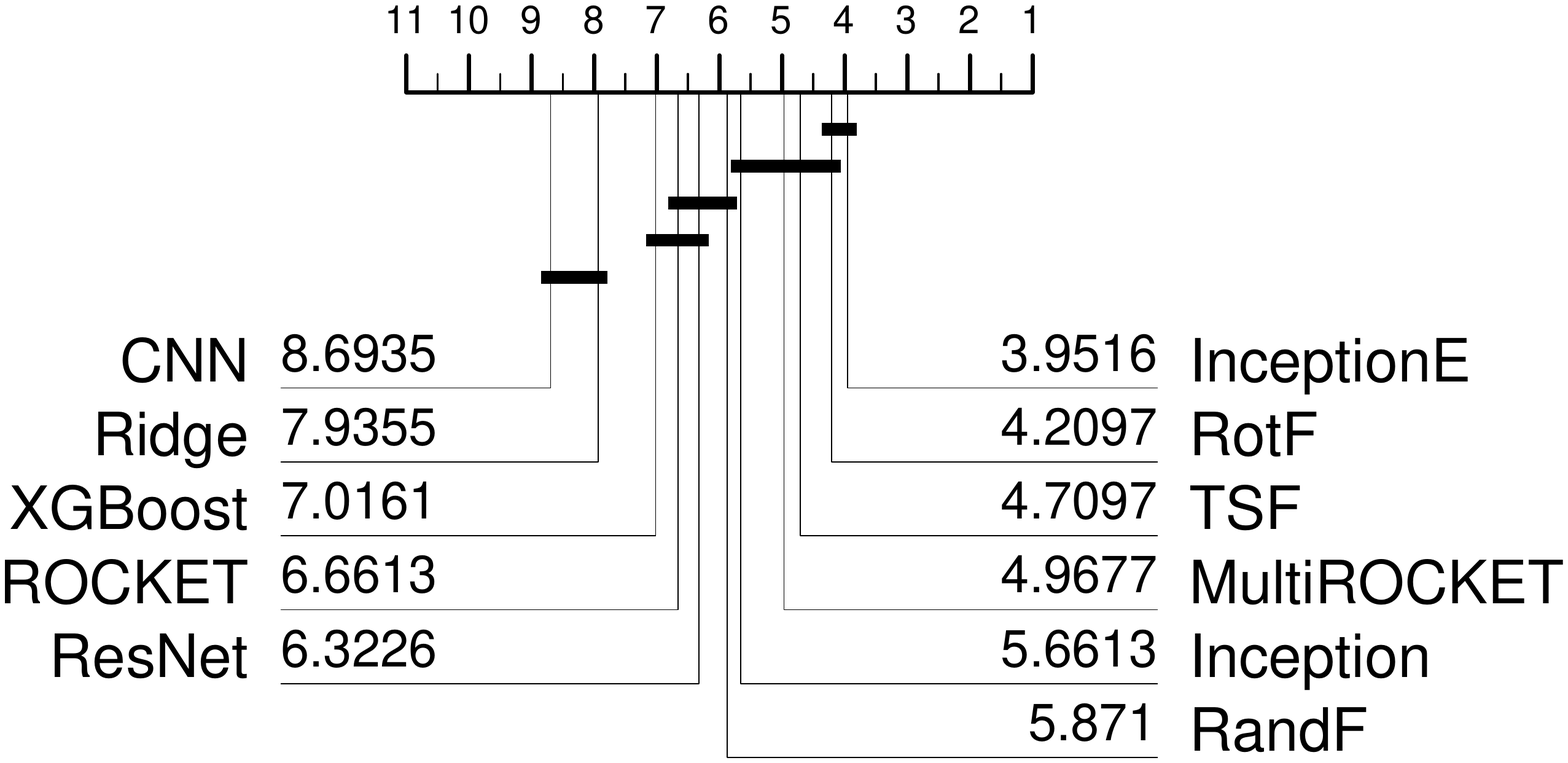}
    \caption{RMSE ranks for 11 regressors on 62 TSER datasets.}
    \label{fig:interim}
\end{figure}

For the next set of experiments, we take the top five algorithms in Figure~\ref{fig:recreate2} and compare these to some alternative adaptations of time series specific algorithms. 
The good performance of XGBoost and RandF suggests we should not overlook standard classifiers. Rotation Forest (RotF)~\cite{rodriguez2006rotation} is a classifier that can be easily adapted to regression by simple averaging~\cite{pardo2013rotation}. It has been shown to be particularly effective for problems with all real valued attributes, including time series~\cite{bagnall2018rotation, bagnall2017great}. Hence, we include a regression adaption in this round of experiments. We also add in the standard Ridge regressor for completeness sake. Besides, the open source toolkit aeon\footnote{https://github.com/aeon-toolkit/aeon} includes two regression implementations not previously evaluated in the context of TSER. TimeSeriesForestRegressor (TSF) is an adaptation of the Time Series Forest classifier~\cite{deng2013time} and CNNRegressor (CNN) is Convolutional Neural Network based on the version described in~\cite{zhao2017cnn}. On further investigation, we found that the results for InceptionTime in~\cite{tan2021time} were created with a single InceptionTime model (Inception). However, in the original work~\cite{ismail2020inceptiontime}, the results supporting InceptionTime as a classifier are found with an ensemble of five InceptionTime models. We include an InceptionTime ensemble model for regression (InceptionE). Furthermore, an improved version of the ROCKET algorithm has been recently publish, known as MultiROCKET \cite{tan22multirocket}. We adapted it to the TSER paradigm accordingly. We provide implementations of MultiROCKET, RotF, and InceptionE in the associated repository. Figure~\ref{fig:interim} shows the results of the three best algorithms from experiments presented in Figure~\ref{fig:recreate2} and six new regressors. We have had to exclude the AustralianRainfall dataset and hence reduce the number of datasets in our study to 62 because of MultiROCKET. The MultiROCKET approach requires over 600GB memory for this dataset and takes more than 15 days to complete. 
The CNN and Ridge regressors are not competitive with the other nine algorithms. InceptionE is significantly better than a single Inception network. As expected, MultiROCKET is significantly better than ROCKET. However, the standout result is that RotF is one of the top performing algorithms along with InceptionE. RotF is the best performing standard algorithm for TSC~\cite{bagnall2017great}, so this is perhaps not surprising. Nevertheless, the fact that an algorithm for standard regression outperforms all the deep learning and time series specific approaches but InceptionE is indicative of the scope for improvement in the field of TSER.

\subsection{Evaluation of new TSER algorithms}
\label{sec:new}
\begin{figure}[htb]
    \centering
\includegraphics[width=0.475\textwidth,trim={1cm 5.5cm 0.5cm 5cm},clip]{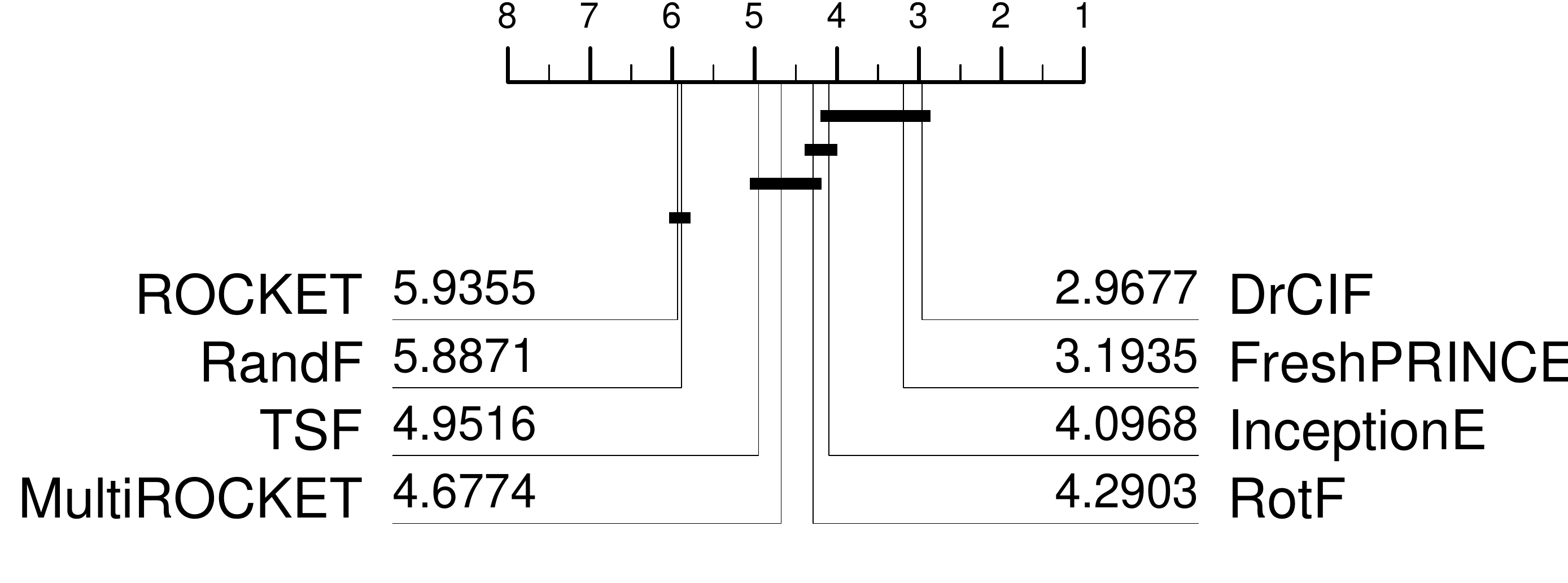} 
    \caption{RMSE ranks for two feature based regressors, DrCIF and freshPRINCE, with the six best performing regressors on 62 TSER datasets.}
    \label{fig:good}
\end{figure}
\begin{figure}[htb]
    \centering
\includegraphics[width=0.475\textwidth,trim={0cm 0cm 0cm 0cm},clip]{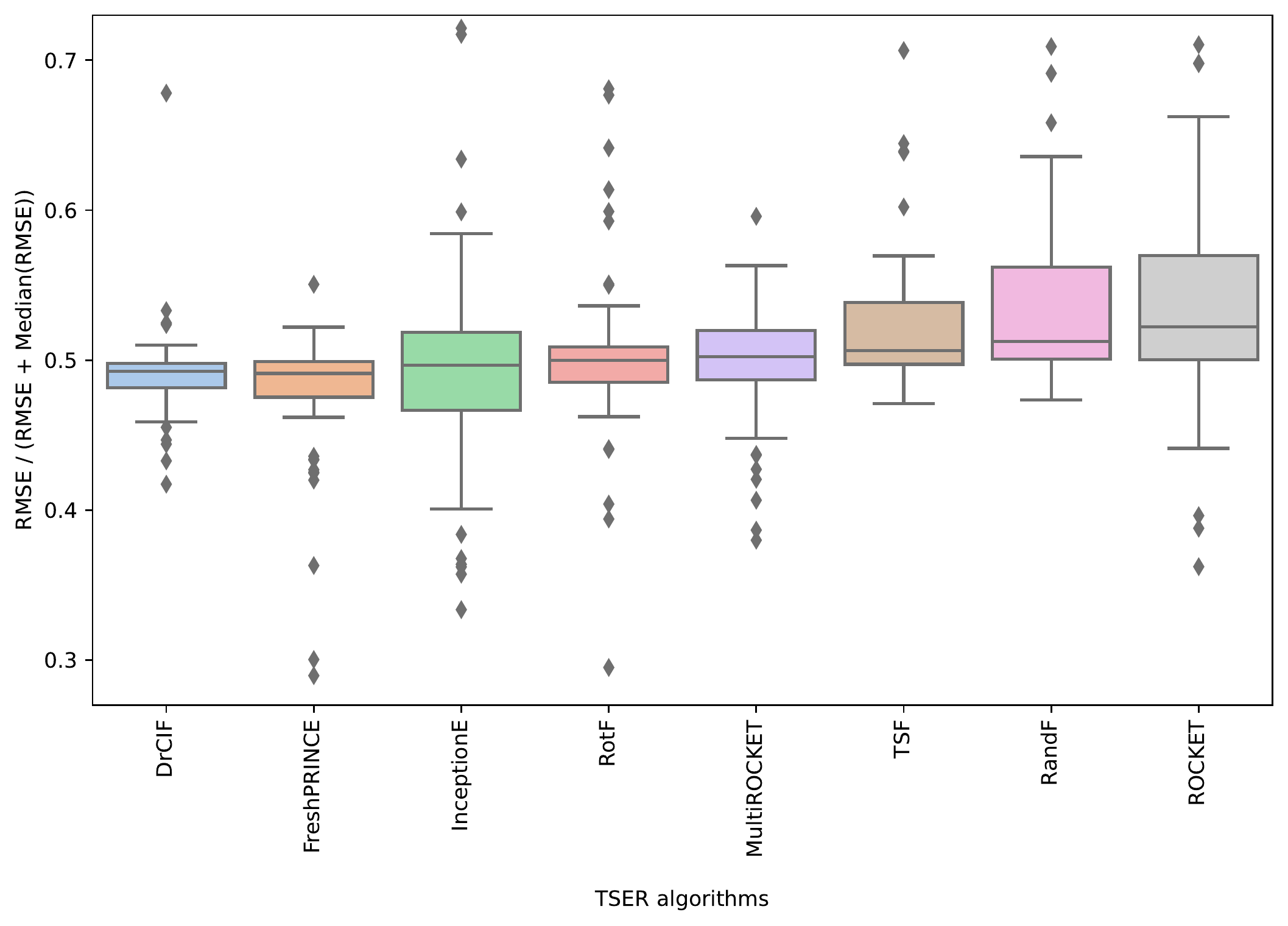} 
    \caption{Distribution of relative RMSE for eight regressors (lower values are better).} 
    \label{fig:boxplot}
\end{figure}
\begin{figure*}[htb]
    \centering
\includegraphics[width=\textwidth,trim={0cm 0cm 0cm 0cm},clip]{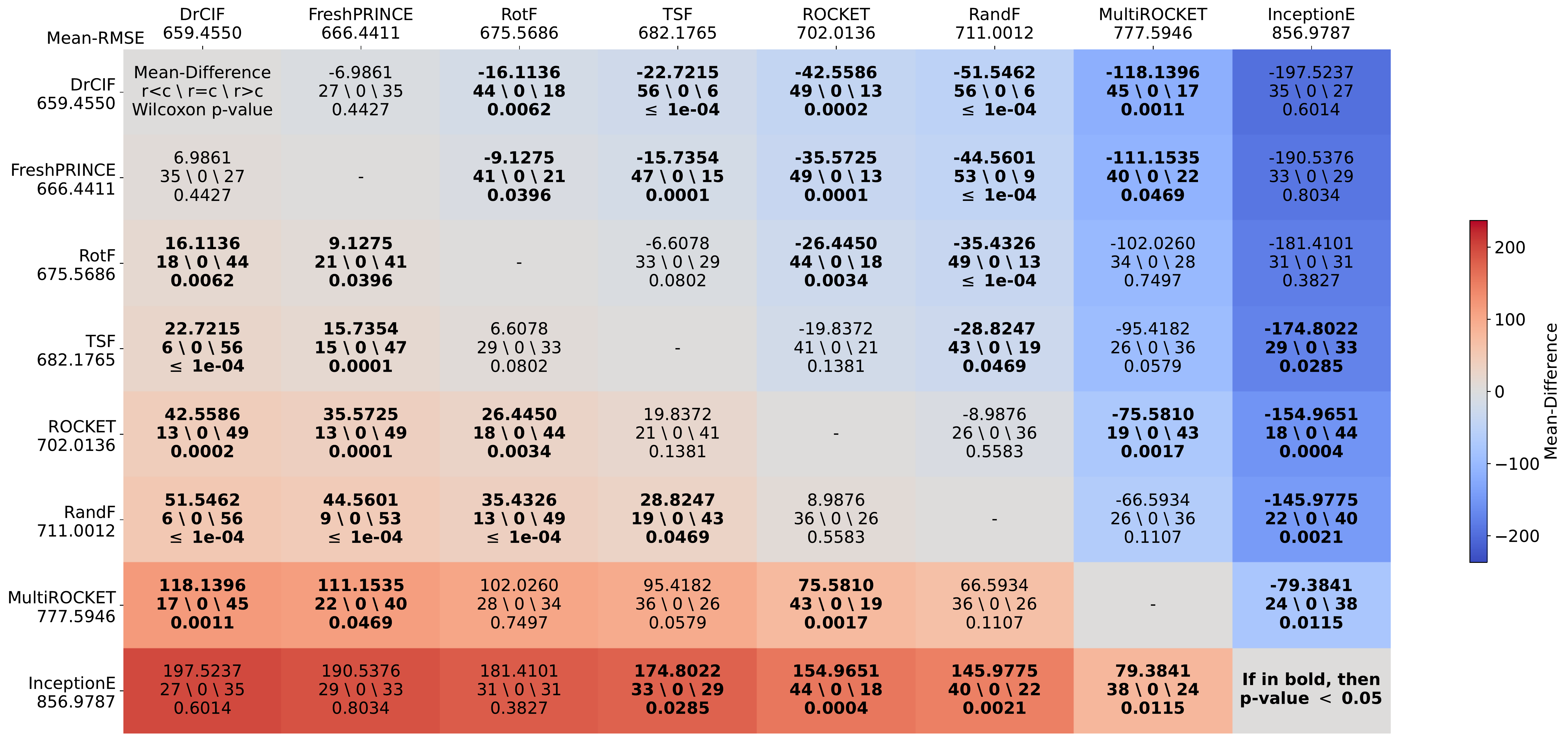} 
    \caption{Summary performance results for the best eight regressors.} 
    \label{fig:heatmap}
\end{figure*}

We now compare the top six algorithms from the previous experiment with our two regressors based on unsupervised feature extraction, DrCIF and FreshPRINCE. Note that the single Inception network is not included as InceptionE always outperforms it. Moreover, we also decided to include ROCKET as it was the top performing approach in the original study of \cite{tan2021time}. Figure~\ref{fig:good} shows that both of the new regressors are in the top clique, and are significantly better in terms of averaged rank for RMSE than RotF, which is the top performing standard approach of all others we have tried. The InceptionE is the third best algorithm. InceptionE is often very good: it is top ranked on 18 of the 62 problems. However, it also fails spectacularly on many problems. Figure~\ref{fig:boxplot} shows the boxplot for the relative deviation of the RMSE over all problems. A figure below 0.5 means the algorithm is better than the median RMSE for that particular problem. We observe that overall, only DrCIF and FreshPRINCE are consistently better than the median performance, and the distribution is tightly coupled. InceptionE has the widest spread.
Figure~\ref{fig:heatmap} elaborates on the analysis of the results. In this case, a heatmap with the average RMSE results is shown, done with a novel tool\footnote{\url{https://github.com/MSD-IRIMAS/Multi_Comparison_Matrix}}. As can be seen, the average results obtained by DrCIF and FreshPRINCE are much better than the rest. Finally, as previously observed, InceptionE shows the worst average RMSE results, which indicates that it is not a very stable method, as its rank is the 3rd, but in terms of mean RMSE it is the 8th.

\section{Analysis} \label{sec:analysis}

We explore our results in more detail to better profile the regressors and gain insights into the drivers behind their performance.

\subsection{Run time}
\begin{figure}[htb]
    \centering
\includegraphics[width=0.475\textwidth,trim={2cm 3.5cm 2cm 3.5cm},clip]{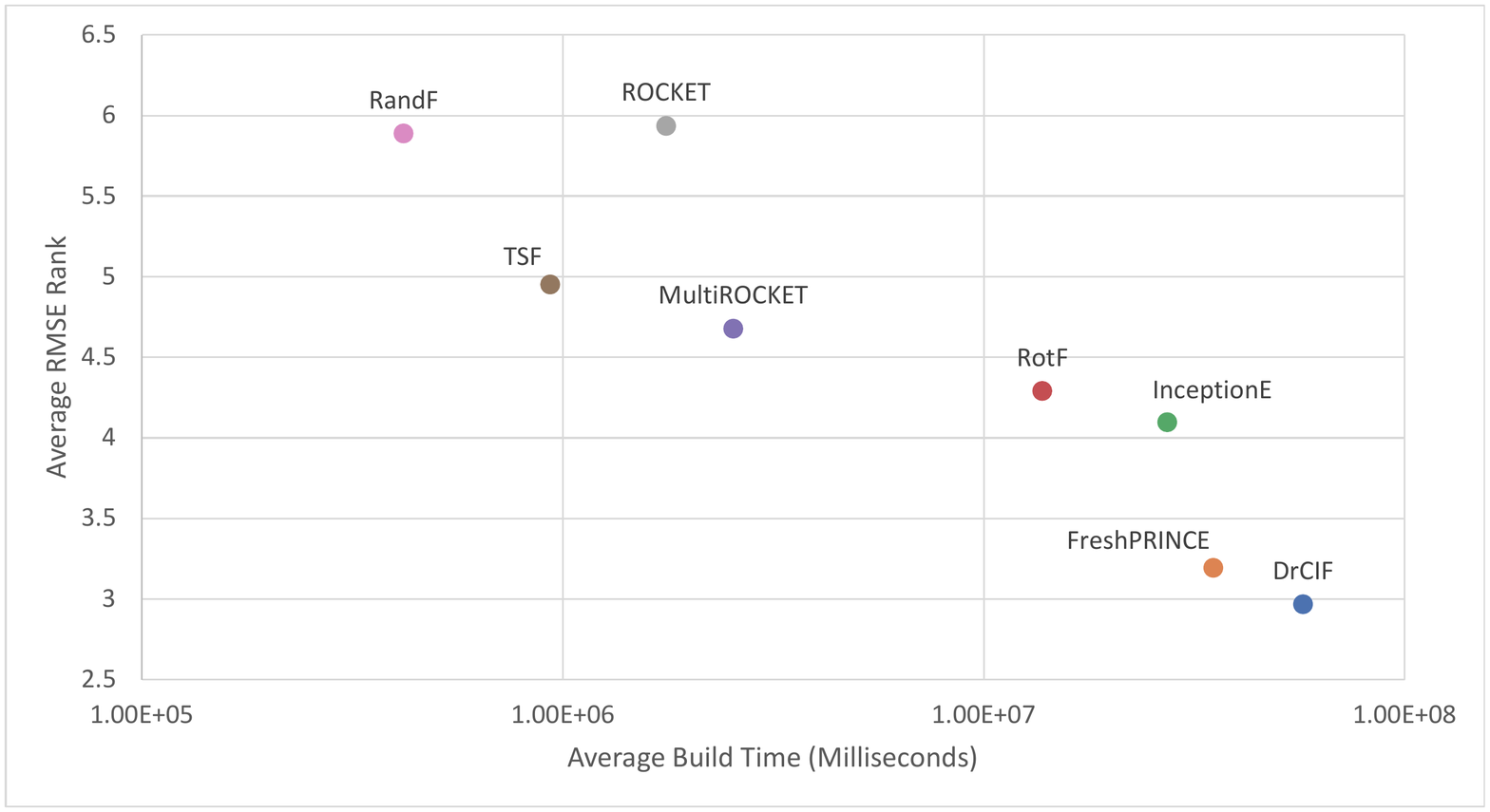} 
    \caption{Run time in milliseconds (log scale average over all datasets) plotted against average rank for RMSE.} 
    \label{fig:runtime}
\end{figure}

Figure~\ref{fig:runtime} shows the average rank RMSE against the run time (on a log scale) for the eight regressors from Figure~\ref{fig:boxplot}. We see a direct trade off between runtime and performance. All algorithms run on a single thread CPU except for InceptionE, which ran on a GPU. This means the graph is very flattering for InceptionE. Even on a GPU it is slower than RotF on a CPU.

\subsection{Performance by Data Characteristics}
We break down the performance of regressors by the core characteristics of the data to help gain insights into when different algorithms perform well. We stress this is purely exploratory: the relatively small number of datasets in each category, in brackets, preclude useful significant testing.
\begin{table}[htb]
	\caption{Average rank RMSE split by number of train cases.}
	\centering
	\small
    \resizebox{\linewidth}{!}{
	\begin{tabular}{l|c|c|c|c}
 \toprule \toprule
		            & $<$300 (15) & 300-999 (23) & 1000-5000 (13) & $>$5000 (11)     \\ \midrule
        DrCIF & \textbf{2.87} & 3.39 & \textit{2.85} & 3.91 \\ 
        FreshPRINCE & \textit{3.60} & \textbf{2.87} & 2.92 & \textbf{2.27} \\ 
        InceptionE & 3.87 & \textit{3.17} & \textbf{2.69} & 3.91 \\ 
        RotF & 4.67 & 4.57 & 5.46 & 4.73 \\ 
        MultiROCKET & 4.27 & 4.30 & 3.85 & \textit{3.73} \\ 
        TSF & 5.13 & 5.48 & 5.08 & 5.55 \\ 
        RandF & 6.07 & 6.04 & 4.31 & 4.36 \\ 
        ROCKET & 5.53 & 6.17 & 5.69 & 5.55 \\ 
  \bottomrule \bottomrule
	\end{tabular}
 }
	\label{tab:bytrain}
\end{table}
Table~\ref{tab:bytrain} shows the average rank RMSE when we group problems by the number of training cases. The pattern is that DrCIF is better with a small number of cases, FreshPRINCE is better with larger train set size. 
\begin{table}[htb]
    \caption{Average rank RMSE split by number of channels.}
    \centering
    \small
    \begin{tabular}{l|c|c|c|c}
    \toprule \toprule
            & 1 (28) & 2 (9) & 3 or 4 (14) & $>$ 5 (11)  \\ \midrule
        DrCIF & \textit{3.29} & \textbf{2.90} & \textbf{2.33} & \textit{2.91} \\ 
        FreshPRINCE & \textbf{2.96} & \textit{3.80} & 4.00 & \textbf{2.55} \\ 
        InceptionE & 4.21 & \textbf{2.90} & 3.58 & 5.09 \\ 
        RotF & 4.25 & 5.50 & 4.58 & \textit{2.91} \\ 
        MultiROCKET & 4.04 & 4.40 & 5.75 & 5.45 \\ 
        TSF & 5.50 & 5.10 & \textit{3.33} & 5.45 \\ 
        RandF & 6.21 & 5.70 & 5.92 & 5.18 \\ 
        ROCKET & 5.54 & 5.70 & 6.50 & 6.45 \\        \bottomrule \bottomrule
    \end{tabular}
    \label{tab:bydimension}
\end{table}
Table~\ref{tab:bydimension} shows the average rank RMSE when we group problems by the number of channels. DrCIF performs relatively better than FreshPRINCE on multivariate problems with 2, 3 or 4 channels. Whereas FreshPRINCE achieves better results when dealing with univariate and multivariate datasets with more than 5 channels.  
\begin{table}[htb]
    \caption{Average rank RMSE split by series length (there are no problems with length 366-999).}
    \centering
    \small
    \resizebox{\linewidth}{!}{
    \begin{tabular}{l|c|c|c|c}
    \toprule \toprule
         & $<$50 (14) & 50-150 (13) & 151-365 (16) & $>$1000 (19)\\  \midrule
        DrCIF & \textbf{2.69} & \textbf{2.00} & \textbf{2.64} & 3.95 \\ 
        FreshPRINCE & 3.77 & 4.25 & \textit{3.00} & \textbf{2.58} \\ 
        InceptionE & 4.23 & 5.63 & 4.64 & \textit{2.74} \\ 
        RotF & \textit{3.69} & 4.13 & 4.05 & 5.05 \\ 
        MultiROCKET & 5.15 & \textit{3.88} & 5.23 & 4.05 \\ 
        TSF & 4.85 & 4.63 & 4.41 & 5.79 \\ 
        RandF & 5.08 & 5.88 & 6.09 & 6.21 \\ 
        ROCKET & 6.54 & 5.63 & 5.95 & 5.63 \\ 
        \bottomrule \bottomrule
    \end{tabular}
    }
    \label{tab:bylength}
\end{table}
Table~\ref{tab:bylength} shows the average rank RMSE when we group problems by the series length. Surprisingly, the interval based DrCIF also performs relatively better than FreshPRINCE with short series but worse with longer series (length $>$1000). 

Finally, we also assessed relative performance for different problem types but did not detect any interesting trends.

\subsection{Ablation of FreshPRINCE}
\begin{figure}[htb]
    \centering
\includegraphics[width=0.475\textwidth,trim={0.7cm 9.2cm 0.5cm 5cm},clip]{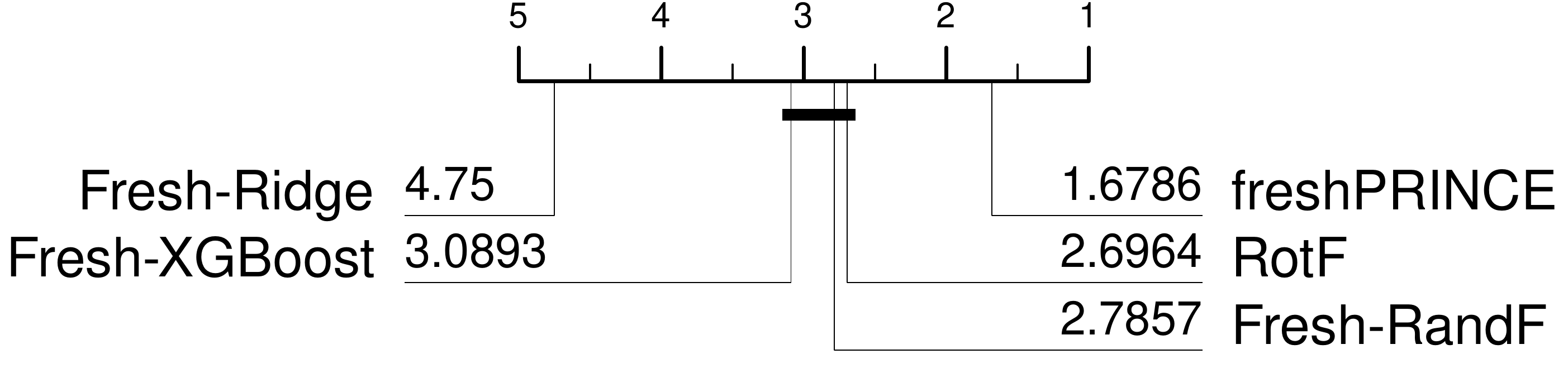} 
    \caption{RMSE ranks for freshPRINCE, standard RotF and TSFresh (Fresh) with alternative regressors.} 
    \label{fig:fp}
\end{figure}

FreshPRINCE is a pipeline of a TSFresh transform and a RotF regressor. We address the question of whether the performance of this regressor is due to the transform, the regressor or both. Figure~\ref{fig:fp} summarises the performance of freshPRINCE, RotF on the raw series and TSFresh transform followed by an alternative regressor. It demonstrates that transforming followed by RandF or XGBoost are no better than simply applying RotF to the raw data. We conclude that it is the combination of transform and regressor that give significantly better performance.
\begin{figure}[htb]
    \centering
\includegraphics[width=0.45\textwidth,trim={0cm 0cm 0cm 0cm},clip]{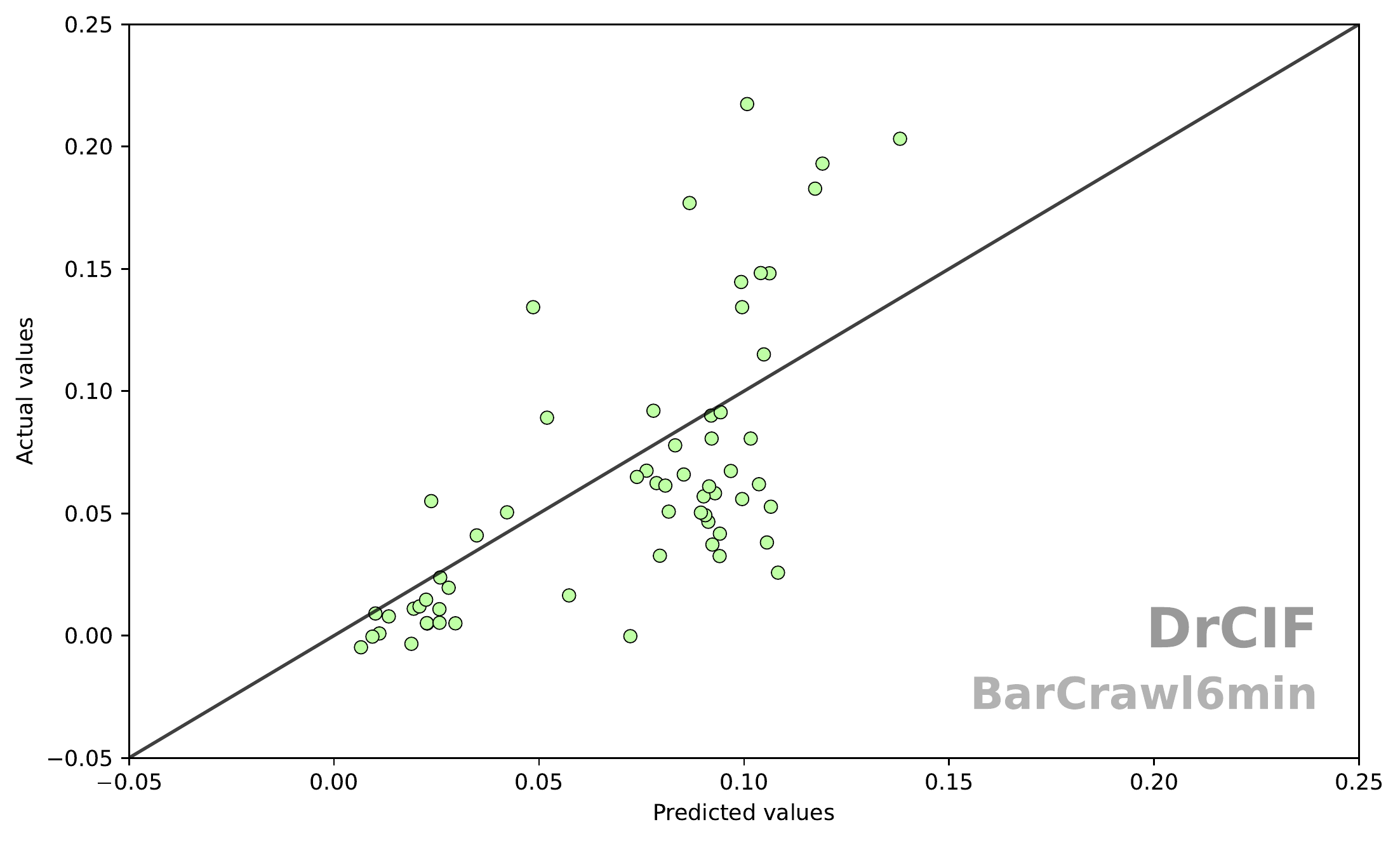} 
    \caption{Scatter plot of predicted vs actual for DrCIF on BarCrawl6min.} 
    \label{fig:bar1}
\end{figure}
\begin{figure}[htb]
    \centering
\includegraphics[width=0.45\textwidth,trim={0cm 0cm 0cm 0cm},clip]{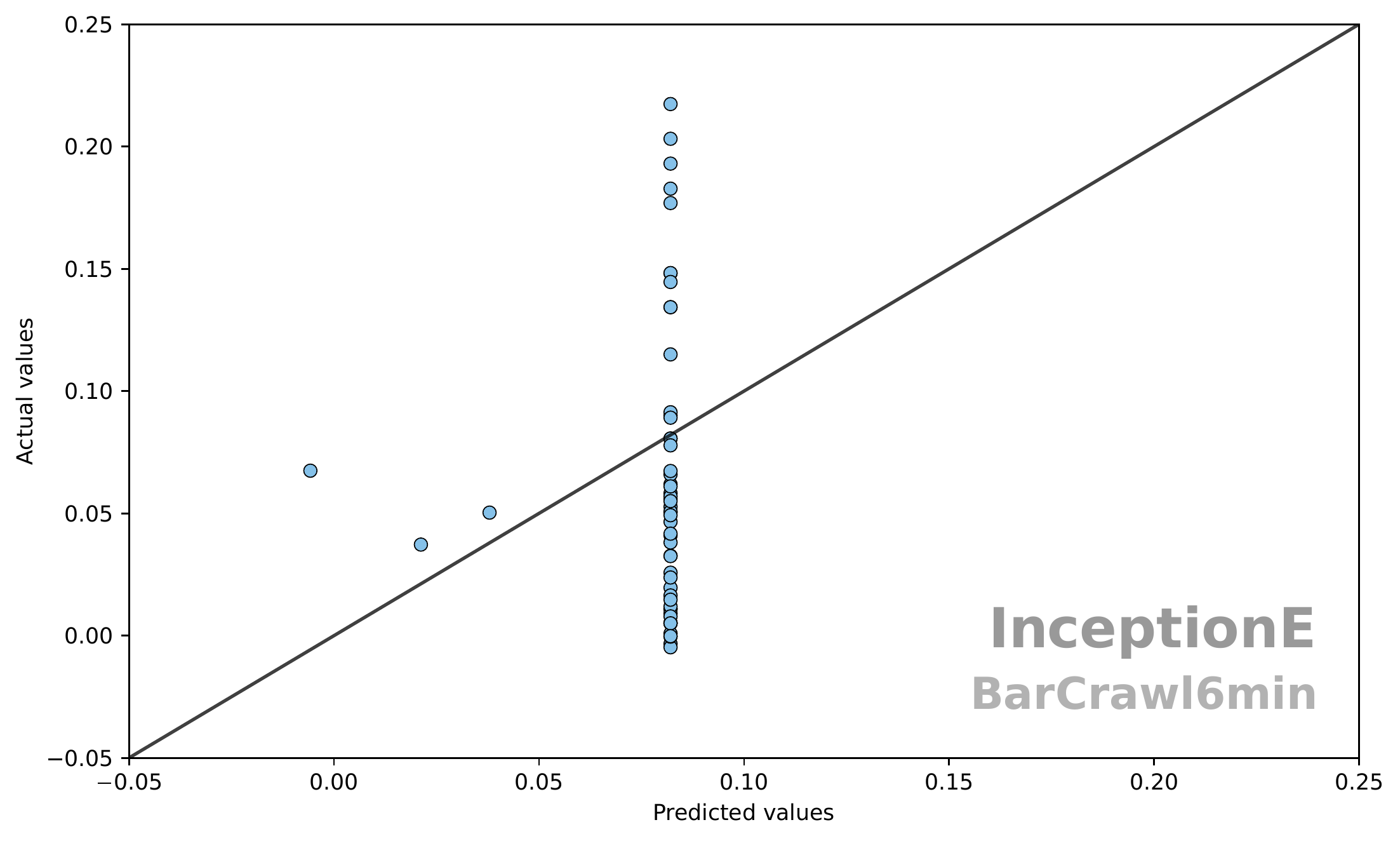} 
    \caption{Scatter plot of predicted vs actual for InceptionE on BarCrawl6min.} 
    \label{fig:bar2}
\end{figure}
There is very little agreement between InceptionE and the other regressors. 
We believe one reason InceptionE performance is so variable is it sometimes completely fails to find anything useful in a dataset where other models have at least some predictive power. To demonstrate this, we look at the standardised residuals of DrCIF and InceptionE on the BarCrawl6min dataset. The time series are accelerometer data, and the response variable is the transdermal alcohol concentration of the test subjects. The response variable is bounded below by zero. 
\begin{figure}[htb]
    \centering
\includegraphics[width=0.45\textwidth,trim={0cm 0cm 0cm 0cm},clip]{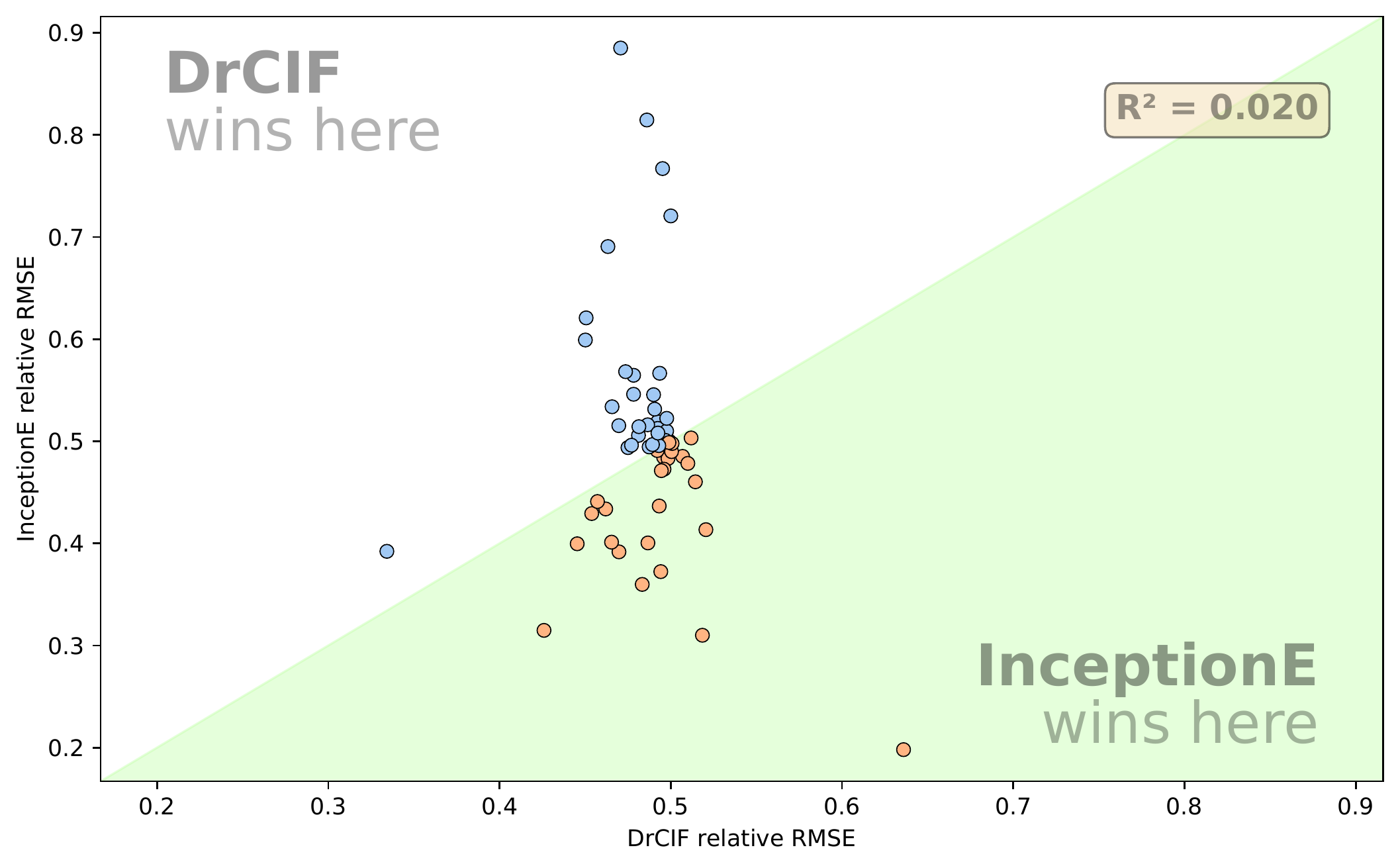} 
    \caption{Scatter plot of relative RMSE for DrCIF and InceptionE.} 
    \label{fig:scatter1}
\end{figure}
In traditional regression, the analyst might look to transform the response with, for example, a Yeo-Johnson transform~\cite{yeo00power}. We are interested in performance over multiple datasets without hand tailored transforms. The RMSE for the default train-test partition for DrCIF is 0.0017 and for InceptionE it is 0.0045. If we plot the predicted response vs the actual response for DrCIF (Figure~\ref{fig:bar1}) we see that DrCIF is making negative predictions for low actual values, and underestimating higher values and there seems to be some heteroscedasticity in the residuals. Nevertheless, it has definitely found some relationship between the regressor series and the response. However, if we see the same plot for InceptionE (Figure~\ref{fig:bar2}), we see that InceptionE is nearly always predicting the same value of 0.082. It is likely that careful configuration, tuning and transform of the data may improve InceptionE. However, the same is true for DrCIF and freshPRINCE. We are using InceptionE in the way recommended by its creators~\cite{ismail2020inceptiontime}. 

Furthermore, Figure~\ref{fig:scatter1} shows the scatter plot of the standardised RMSE used in Figure~\ref{fig:boxplot} for DrCIF vs InceptionE. There is no correlation. For context, for our top four regressors, the strongest correlation is between DrCIF and RotF ($R^2 =0.27$) and DrCIF/freshPRINCE are very weakly correlated ($R^2 = 0.13$). This diversity suggests there may be some value in ensembling.

\section{Conclusions} \label{sec:conclusions}
We have proposed new algorithms for Time Series Extrinsic Regression (TSER) based on classifiers and conducted an extensive experimental study. We have increased the TSER archive size from 19 to 63, introduced improved versions of regressors used in the previous study and shown our new adaptations of classification algorithms are significantly better than the best alternatives. There are several limitations to this study. The data is not randomly sampled (we have taken problems from where ever we can find them) and some domains may be over represented; we have not tuned any of these regressors (the computation required to tune these algorithms over 63 datasets would be prohibitive); we have not looked at more complex diagnostics of performance such as residual analysis. Nevertheless, we believe we have made a significant contribution to advance the new field of TSER. RotF outperforms all previously assessed regressors, and DrCIF and FreshPRINCE are the only TSER algorithms so far proposed to significantly outperform all standard regression algorithms. We have made all our experiments reproducible by releasing structured source code compatible with standard toolkits, guidance on reproducing experiments, and all of our results. 

We believe there is scope for further improvement for algorithms for TSER. Adapting supervised Time Series Classification (TSC) approaches may help further leverage of this popular theme for research. InceptionE is the most promising deep learning approach, and perhaps it can be engineered to avoid the catastrophic failure it tends towards with smaller train set sizes. Heterogeneous ensembles are very successful for TSC, and the diversity in performance between the best performing algorithms suggest this may, with careful adaptation, translate to TSER. We will continue to enhance the repository with more problems and would welcome all donations.

\section*{Acknowledgments}
This work has been partially subsidised by ``Agencia Española de Investigación (España)'' (grant ref.: PID2020-115454GB-C22 / AEI / 10.13039 / 501100011033). David Guijo-Rubio’s research has been subsidised by the University of Córdoba through grants to Public Universities for the requalification of the Spanish university system of the Ministry of Universities, financed by the European Union - NextGenerationEU (grant reference: UCOR01MS). It is also supported by EPSRC (grant reference EP/W030756/1). Guilherme Arcencio's research has been subsidised by the São Paulo Research Foundation (FAPESP) (grants references: \#2022/12486-4 and \#2022/00305-5). Diego Furtado Silva's research has also been subsidised by FAPESP (grant reference: \#2022/03176-1). Some of the experiments were carried out on the High Performance Computing Cluster supported by the Research and Specialist Computing Support service at the University of East Anglia. We would like to thank all those responsible for helping maintain the time series classification archives and those contributing to open source implementations of the algorithms.



\appendices

\section{Data description} \label{sec:data}

The list of all 63 datasets in the archive is shown in Table \ref{tab:datasets}. More details on the 44 datasets we have added are given in this section. 

\begin{table*}[ht!]
\centering
    \caption{Time series datasets in the TSER archive.}
    \label{tab:datasets}
    
    \resizebox{0.8\textwidth}{!}{
    \begin{tabular}{l|cccccc}
        \toprule \toprule
        Dataset & Train size & Test size & Length & $\#$ Dims & Missing & Category \\
        \midrule
        AppliancesEnergy & 95 & 42 & 144 & 24 & No & Energy monitoring\\
        HouseholdPowerConsumption1 & 745 & 686 & 1440 & 5 & Yes & Energy monitoring\\
        HouseholdPowerConsumption2 & 745 & 686 & 1440 & 5 & Yes & Energy monitoring\\
        AustraliaRainfall & 112186 & 48081 & 24 & 3 & No & Environment monitoring\\
        BeijingPM10Quality & 11918 & 5048 & 24 & 9 & Yes & Environment monitoring\\
        BeijingPM25Quality & 11918 & 5048 & 24 & 9 & Yes & Environment monitoring\\
        BenzeneConcentration & 3349 & 5163 & 240 & 8 & Yes & Environment monitoring\\
        FloodModeling1 & 471 & 202 & 266 & 1 & No & Environment monitoring\\
        FloodModeling2 & 466 & 201 & 266 & 1 & No & Environment monitoring\\
        FloodModeling3 & 429 & 184 & 266 & 1 & No & Environment monitoring\\
        LiveFuelMoistureContent & 3493 & 1510 & 365 & 7 & No & Environment monitoring\\
        BIDMC32HR & 5550 & 2399 & 4000 & 2 & No & Health monitoring\\
        BIDMC32RR & 5471 & 2399 & 4000 & 2 & No & Health monitoring\\
        BIDMC32SpO2 & 5550 & 2399 & 4000 & 2 & No & Health monitoring\\
        Covid3Month & 140 & 61 & 84 & 1 & No & Health monitoring\\
        IEEEPPG & 1768 & 1328 & 1000 & 5 & No & Health monitoring\\
        PPGDalia & 43215 & 21482 & 256-512$^D$ & 4 & No & Health monitoring\\
        NewsHeadlineSentiment & 58213 & 24951 & 144 & 3 & No & Sentiment analysis\\
        NewsTitleSentiment & 58213 & 24951 & 144 & 3 & No & Sentiment analysis\\
        \midrule
        DailyOilGasPrices & 133 & 58 & 30 & 2 & No & Economic analysis\\
        ChilledWaterPredictor & 321 & 138 & 168 & 4 & No & Energy monitoring\\
        ElectricityPredictor & 567 & 243 & 168 & 4 & No & Energy monitoring\\
        HotwaterPredictor & 245 & 106 & 168 & 4 & No & Energy monitoring\\
        OccupancyDetectionLight & 237 & 103 & 60 & 3 & No & Energy monitoring\\
        SolarRadiationAndalusia & 470 & 202 & 365 & 2 & Yes & Energy monitoring\\
        SteamPredictor & 210 & 90 & 168 & 4 & No & Energy monitoring\\
        TetuanEnergyConsumption & 254 & 110 & 144 & 5 & No & Energy monitoring\\
        WindTurbinePower & 596 & 256 & 144 & 1 & No & Energy monitoring\\
        AcousticContaminationMadrid & 166 & 72 & 365 & 1 & Yes & Environment monitoring\\
        AluminiumConcentration & 440 & 189 & 2542 & 1 & No & Environment monitoring\\
        BeijingIntAirportPM25Quality & 1099 & 472 & 24 & 6 & No & Environment monitoring\\
        BoronConcentration & 438 & 188 & 2542 & 1 & No & Environment monitoring\\
        CalciumConcentration & 444 & 191 & 2307 & 1 & No & Environment monitoring\\
        CopperConcentration & 440 & 189 & 2542 & 1 & No & Environment monitoring\\
        DailyTemperatureLatitude & 27440 & 11760 & 365 & 1 & No & Environment monitoring\\
        DhakaHourlyAirQuality & 1447 & 621 & 24 & 1 & No & Environment monitoring\\
        IronConcentration & 427 & 184 & 1716 & 1 & No & Environment monitoring\\
        MadridPM10Quality & 4845 & 2077 & 168 & 3 & Yes & Environment monitoring\\
        MagnesiumConcentration & 1560 & 669 & 3578 & 1 & No & Environment monitoring\\
        ManganeseConcentration & 427 & 184 & 1716 & 1 & No & Environment monitoring\\
        MetroInterstateTrafficVolume & 849 & 365 & 24 & 4 & No & Environment monitoring\\
        ParkingBirmingham & 1391 & 597 & 14-18$^S$ & 1 & No & Environment monitoring\\
        PhosphorusConcentration & 1573 & 675 & 3578 & 1 & No & Environment monitoring\\
        PotassiumConcentration & 1561 & 669 & 3578 & 1 & No & Environment monitoring\\
        PrecipitationAndalusia & 470 & 202 & 365 & 4 & Yes & Environment monitoring\\
        SierraNevadaMountainsSnow & 350 & 150 & 30 & 3 & No & Environment monitoring\\
        SodiumConcentration & 424 & 183 & 1716 & 1 & No & Environment monitoring\\
        SulphurConcentration & 444 & 191 & 2307 & 1 & No & Environment monitoring\\
        ZincConcentration & 445 & 191 & 2307 & 1 & No & Environment monitoring\\
        ElectricMotorTemperature & 15503 & 6645 & 60 & 6 & No & Equipment monitoring\\
        GasSensorArrayAcetone & 324 & 140 & 7500 & 1 & No & Equipment monitoring\\
        GasSensorArrayEthanol & 324 & 140 & 7500 & 1 & No & Equipment monitoring\\
        LPGasMonitoringHomeActivity & 2017 & 865 & 100 & 1 & No & Equipment monitoring\\
        MethaneMonitoringHomeActivity & 2017 & 865 & 100 & 1 & No & Equipment monitoring\\
        WaveDataTension & 1325 & 568 & 57 & 1 & No & Equipment monitoring\\
        BarCrawl6min & 140 & 61 & 360 & 3 & No & Health monitoring\\
        Covid19Andalusia & 142 & 62 & 91 & 1 & No & Health monitoring\\
        VentilatorPressure & 52815 & 22635 & 80 & 2 & No & Health monitoring\\
        BinanceCoinSentiment & 184 & 79 & 24 & 2 & No & Sentiment analysis\\
        BitcoinSentiment & 232 & 100 & 24 & 2 & No & Sentiment analysis\\
        CardanoSentiment & 74 & 33 & 24 & 2 & No & Sentiment analysis\\
        EthereumSentiment & 249 & 107 & 24 & 2 & No & Sentiment analysis\\
        NaturalGasPricesSentiment & 65 & 28 & 20 & 1 & No & Sentiment analysis\\
        \bottomrule \bottomrule
        \multicolumn{7}{l}{For those datasets with unequal-length time series: $^S$ indicates that the unequal-length time series are with respect to}\\
        \multicolumn{7}{l}{the samples, whereas $^D$ indicates that is with respect to the dimensions.}\\
    \end{tabular}
    }
\end{table*}

\subsection{Economic analysis}

\subsubsection{Oil and natural gas prices}

A dataset published on Kaggle\textsuperscript{\ref{dailyOilGasPrices}} consists of historical prices of Brent Oil, Crude Oil WTI, Natural Gas, and Heating Oil from 2000 to 2022. We created the \textbf{DailyOilGasPrices} by using 30 consecutive business days of Crude Oil WTI close prices and traded volumes as predictors and the average natural gas close price during each 30-day time frame as the target variable. The final dataset has 191 2-dimensional time series of length 30, of which 70\% were randomly sampled as training data and the remaining 30\% as testing data. This type of model could help companies and governments to better analyse and predict economic situations and correlations regarding oil and natural gas.

\subsection{Energy monitoring}

\subsubsection{ASHRAE - Great Energy Predictor III}

This dataset, published on Kaggle\textsuperscript{\ref{EnergyBuilding}}, aims to assess the value of energy efficiency improvements. For that purpose, four types of sources are identified: electricity, chilled water, steam and hot water. The goal is to estimate the energy consumption in kWh.

Dimensions correspond to the air temperature, dew temperature, wind direction and wind speed. These values were taken hourly during a week, and the output is the meter reading of the four aforementioned sources. In this way, we created four datasets: \textbf{ChilledWaterPredictor}, \textbf{ElectricityPredictor}, \textbf{HotwaterPredictor}, and \textbf{SteamPredictor}. Each dataset has a different number of time series as they correspond to different buildings using those sources. In this sense, \textbf{ChilledWaterPredictor} resulted in 459 4-dimensional time series of length 168, \textbf{ElectricityPredictor} resulted in 810 4-dimensional time series of length 168, \textbf{HotwaterPredictor} resulted in 351 4-dimensional time series of length 168, and \textbf{SteamPredictor} resulted in 300 4-dimensional time series of length 168. We randomly sampled 70\% of those time series to use as train data and the remaining 30\% as test data.

Even though there is a kaggle post indicating that there is one building with meter reading in kBTU they have been transformed into kWh accordingly.

\subsubsection{Occupancy detection}

In a study by Candanedo and Feldheim~\cite{datasets_CANDANEDO201628}, measurements of temperature, light, $\textrm{CO}_2$, and humidity, collected every minute, were used to detect whether an office room was occupied or not. This data has been made available in the UCI Machine Learning repository\textsuperscript{\ref{OccupancyDetectionLight}}.

We created the \textbf{OccupancyDetectionLight} dataset by reformulating the problem. We used one hour of temperature (in °C), humidity ratio, and $\textrm{CO}_2$ concentration (in ppm) as predictors and the average light during that hour (in Lux) as the response variable. This resulted in 340 3-dimensional time series of length 60. We randomly sampled 70\% of those time series to use as train data and the remaining 30\% as test data. Better models for this data can lead to improvements in energy consumption analysis and prediction.

\hyperfootnotetext{\label{OccupancyDetectionLight}\textsuperscript{\ref*{OccupancyDetectionLight}}\url{https://perma.cc/5ER3-MXG5}}

\subsubsection{Solar radiation in Andalusia}

This dataset has been obtained from the Andalusia Government (Spain)\textsuperscript{\ref{SolarRadiationAndalusia}}. Data was retrieved from different stations of the 8 districts of Andalusia: Almeria, Cadiz, Cordoba, Granada, Huelva, Jaen, Malaga, Sevilla, from 2000 until february 2014. The dataset is known as \textbf{SolarRadiationAndalusia}. Dimensions correspond to daily mean of humidity and temperature, whereas the output is the solar radiation for the same day. As time series take daily values during complete years, data from 2014 is not used. 

The final dataset includes 672 time series with 2 dimensions with length of 365. The training dataset includes randomly samples 70\%, whereas the remaining 30\% forms the testing dataset. 

\subsubsection{Energy consumption in Tetuan}

This dataset, published on UCI Machine Learning Repository\textsuperscript{\ref{TetuanEnergyConsumption}}, aims to estimate the power consumption in three zones in Tetouan~\cite{salam2018comparison}. The new dataset is known as \textbf{TetuanEnergyConsumption}. Data has been collected on a ten minute basis. Hence, time series have 144 values (6 values per hour). A total of 5 dimensions have been identified: temperature, humidity, wind speed, general diffuse flows and diffuse flows. The goal is to estimate the daily average power consumption in the three zones of Tetouan.

The aforementioned dataset includes 364 5-dimensional time series of length 144. 70\% of those 364 time series have been randomly selected for the training set, whereas the remaining 30\% belong to the testing set.

\hyperfootnotetext{\label{TetuanEnergyConsumption}\textsuperscript{\ref*{TetuanEnergyConsumption}}\url{https://perma.cc/NX65-A5B4}}

\subsubsection{Wind turbine power generation}

The ``Wind Turbine Power (kW) Generation Data'' dataset on Kaggle\textsuperscript{\ref{WindTurbinePower}} consists of large amounts of data collected from a wind turbine, from the temperatures at different parts of the turbines, to the angular position of each blade, and the turbine's power output. The measurements were made on a 10-minute basis and span from 2019 to 2021.

We used this data to create the \textbf{WindTurbinePower} dataset. Each time series consists of 144 timepoints, i.e. one day of measurements, and its target variable is the turbine's average power output on that day. From the 76 possible features, we used only the turbine's torque as a predictor, since it was enough to achieve good results and the other features seemed secondary at best. Some instances were removed due to their respective days having fewer than 144 measurements. The resulting dataset has 852 instances, of which 70\% were randomly sampled as training data and the remaining 30\% as testing data.

Good regression models for this dataset should improve the logistics of green energy production, by better predicting the energy output of a wind turbine in a given place and/or season.

\subsection{Environment monitoring}

\subsubsection{Acoustic contamination in Madrid, Spain}

\begin{figure}[htb]
\centering
\includegraphics[width=\columnwidth]{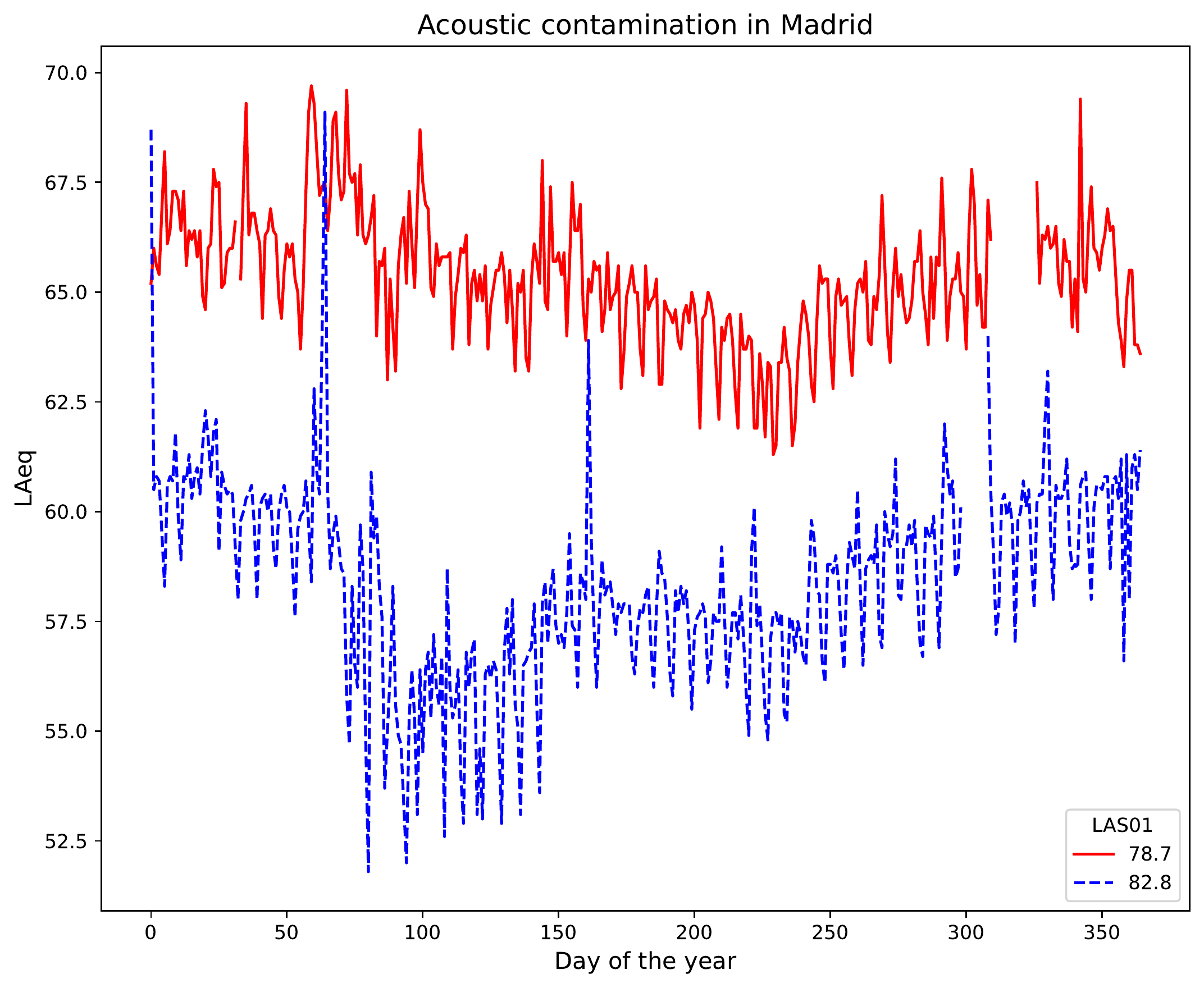}
\caption{Two examples of yearly sound pollution in the AcousticContaminationMadrid dataset}
\label{fig:datasetacoustic}
\end{figure}

This dataset has been made publicly available by the Government of Madrid, Spain\textsuperscript{\ref{AcousticContaminationMadrid}}. Data is collected by a number of stations located in the city of Madrid. This dataset is updated daily since 2014. However, we created the \textbf{AcousticContaminationMadrid} dataset with data up to December 2021. 
The input time series is the LAeq, a fundamental measurement parameter designed to represent a varying sound source over a given time as a single number. Whereas the output time series is the LAS01, the first percentile of sound pressure levels, with A frequency weighting and slow time weighting, recorded during the corresponding period. Examples of such series and outputs are shown in Figure \ref{fig:datasetacoustic}.

The final dataset includes 238 univariate time series. Moreover, their length is 365, which corresponds to daily values taken during a year. The training dataset is composed of randomly selected 70\% of the samples, whereas the remaining 30\% composes the testing dataset.

\subsubsection{Africa soil chemistry}

The Africa Soil Information Service (AfSIS) Soil Chemistry\textsuperscript{\ref{AfricaSoilChemistry}} dataset contains large amounts of dry and wet chemistry data obtained from soil samples collected from many countries throughout Sub-Saharan Africa, from 2009 to 2013. Dry chemistry analysis, such as infrared spectroscopy and X-ray fluorescence, is comparably less expensive than wet chemistry. Therefore, a model which uses only dry chemistry to predict certain nutrients measured by wet chemical analyses is commercially interesting.

The dry chemical measurements were taken using many different machine models, of which we selected four: ``Alpha ZnSe'', ``Alpha KBr'', ``HTSXT'' and ``MPA''. The wet chemistry data for each soil sample includes the quantity of 12 nutrients: Aluminium, Boron, Copper, Iron, Manganese, Sodium, Phosphorous, Potassium, Magnesium, Sulphur, Zync and Calcium. Each dry chemical measurement is a time series where the time axis is wavelength and the y-axis is the respective response.

We paired a third of each dry chemistry machine's experiments to a different nutrient measurement, thus creating 12 datasets. They are \textbf{AluminiumConcentration} (629 cases of length 2542), \textbf{BoronConcentration} (626 cases of length 2542), \textbf{CopperConcentration} (629 cases of length 2542), \textbf{IronConcentration} (611 cases of length 1716), \textbf{ManganeseConcentration} (611 cases of length 1716), \textbf{SodiumConcentration} (607 cases of length 1716), \textbf{PhosphorusConcentration} (2248 cases of length 3578), 
\textbf{PotassiumConcentration} (2230 cases of length 3578), 
\textbf{MagnesiumConcentration} (2229 cases of length 3578), 
\textbf{SulphurConcentration} (635 cases of length 2307), 
\textbf{ZincConcentration} (636 cases of length 2307), and
\textbf{CalciumConcentration} (635 cases of length 2307).

In each dataset, 70\% of cases were randomly sampled as training data and the remaining 30\% as testing data. Three example soil spectrograms are shown in Figure \ref{fig:datasetafricasoil}.

\subsubsection{Beijing Airport PM2.5 contamination}

This dataset was obtained from the UCI Machine Learning Repository\textsuperscript{\ref{BeijingIntAirportPM25Quality}}. The authors \cite{liang2015assessing}, collected hourly data containing the $\textrm{PM}_{2.5}$ data of US Embassy in Beijing, as well as meteorological data from Beijing Capital International Airport. This dataset, known as \textbf{BeijingIntAirportPM25\allowbreak Quality}, includes 6-dimensional time series of 24 points. The dimensions are the dew point, temperature, pressure, combined wind direction, and accumulated hours of snow and rain measured, as mentioned, in the Beijing Capital International Airport. The output is the $\textrm{PM}_{2.5}$ data averaged daily.

The aforementioned dataset includes 1571 6-dimensional time series of length 24. 70\% of those 1571 time series have been randomly selected for the training set, whereas the remaining 30\% belong to the testing set. 

\hyperfootnotetext{\label{BeijingIntAirportPM25Quality}\textsuperscript{\ref*{BeijingIntAirportPM25Quality}}\url{https://perma.cc/G8VS-DKKY}}

\subsubsection{Daily temperature and latitude}

A dataset published on Kaggle\textsuperscript{\ref{DailyTemperatureLatitude}} contains daily temperature data for the 1000 most populous cities in the world, along with their geographic coordinates, from 1980 to 2020.

\begin{figure}[htb]
\centering
\includegraphics[width=\columnwidth]{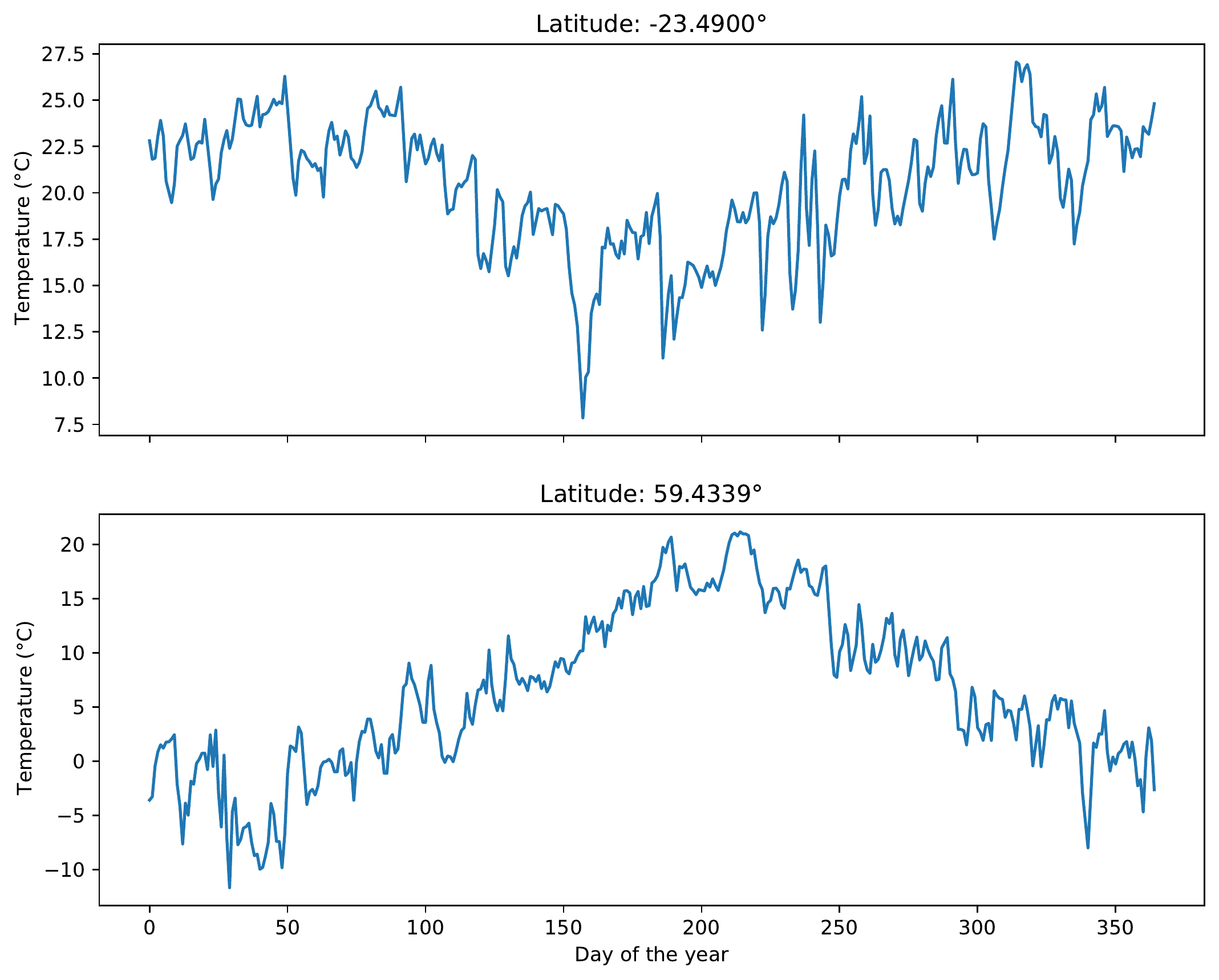}
\caption{Examples of yearly temperature profiles from two different latitudes in the DailyTemperatureLatitude dataset}
\label{fig:datasetlatitude}
\end{figure}

We used this data to create the \textbf{DailyTemperatureLatitude} dataset. We split each city's temperature data into 1 year long time series, i.e. 365 timepoints. Leap years were shortened by averaging the temperatures on the 28\textsuperscript{th} and 29\textsuperscript{th} of February. The predictors are the daily temperatures (in °C) recorded during the year and the response variable is the corresponding city's latitude. The final dataset has 39200 univariate time series of length 365, of which 70\% were randomly sampled as training data and the remaining 30\% as testing data. Two samples of the constructed dataset are shown in Figure \ref{fig:datasetlatitude}.

With exploratory analysis and regression on this dataset, climate change and its effects can be better understood and predicted on a local basis.

\subsubsection{Air quality in Dhaka}

Data sourced from AirNow\textsuperscript{\ref{DhakaHourlyAirQuality2}} and made available on Kaggle\textsuperscript{\ref{DhakaHourlyAirQuality}} comprises 7 years of hourly measurements of fine particulate matter ($\textrm{PM}_{2.5}$) concentrations at the United States Embassy in Dhaka, Bangladesh, along with each corresponding Air Quality Index (AQI).

We used this data to create the \textbf{DhakaHourlyAirQuality} dataset, in which the predictors are 24 hours of $\textrm{PM}_{2.5}$ concentrations and the response variable is the average AQI on that respective day. Thus, better models can be created and tested to more cheaply and quickly evaluate air quality in different cities. The resulting dataset consists of 2068 univariate time series of length 24, of which 70\% were randomly sampled as training data and the remaining 30\% as testing data.

\hyperfootnotetext{\label{DhakaHourlyAirQuality2}\textsuperscript{\ref*{DhakaHourlyAirQuality2}}\url{https://perma.cc/2K2D-LT6T}}

\subsubsection{Madrid PM10 contamination}

This dataset is publicly available on Kaggle\textsuperscript{\ref{MadridPM10Quality}}, even though the data was originally published in the public repository of the Madrid Government\textsuperscript{\ref{MadridPM10Quality2}}. The dataset is known as \textbf{MadridPM10Quality}. Its input data consists in measurements of the level of sulphur dioxide, carbon monoxide and nitric oxide. Time series correspond to hourly values measured during a week. The output value is the weekly averaged $\textrm{PM}_{10}$.

The final dataset includes 6922 3-dimensional time series with a length of 168. The training dataset is composed of randomly selected 70\% of the samples, whereas the remaining 30\% forms the testing dataset.

\hyperfootnotetext{\label{MadridPM10Quality2}\textsuperscript{\ref*{MadridPM10Quality2}}\url{https://perma.cc/5QNM-ZWGS}}

\subsubsection{Metro Interstate Traffic Volume}

This dataset is publicly available on the UCI Machine Learning Repository\textsuperscript{\ref{MetroInterstateTrafficVolume}}. Data consists of hourly traffic volume for MN DoT ATR station 301, roughly midway between Minneapolis and St Paul, collected from 2012 to 2018. The dataset, known as \textbf{MetroInterstateTrafficVolume}, aims to estimate the daily average traffic volume for the aforementioned road. The input dimensions consists of the average temperature, the amount of rain and snow, and the percentage of cloud cover. 

The final dataset includes 1214 time series with 4 dimensions. Moreover, their length is 24, which corresponds to hourly measures taken during a day. The training dataset includes a randomly selected 70\% of the whole data, whereas the remaining 30\% forms the testing dataset.

\subsubsection{Parking Birmingham}

This dataset is publicly available on the UCI Machine Learning Repository\textsuperscript{\ref{ParkingBirmingham}}. The authors \cite{stolfi2017predicting} collected data from car parks in Birmingham (United Kingdom) operated by National Car Parks from Birmingham City Council. The dataset, known as \textbf{ParkingBirmingham}, aims to estimate occupancy rates from 2016/10/04 to 2016/12/19. Input time series is the number of parked cars every hour, whereas the output is the occupancy rate. The total number of hours measured per day varies from 14 to 18.

The aforementioned dataset includes 1988 unequal length time series (with lengths between 14 and 18). 70\% of those 1988 time series have been randomly selected for the training set, whereas the remaining 30\% belong to the testing set.

\hyperfootnotetext{\label{ParkingBirmingham}\textsuperscript{\ref*{ParkingBirmingham}}\url{https://perma.cc/3LAJ-YTSG}}

\subsubsection{Precipitation in Andalusia}

\begin{figure}[htb]
\centering
\includegraphics[width=\columnwidth]{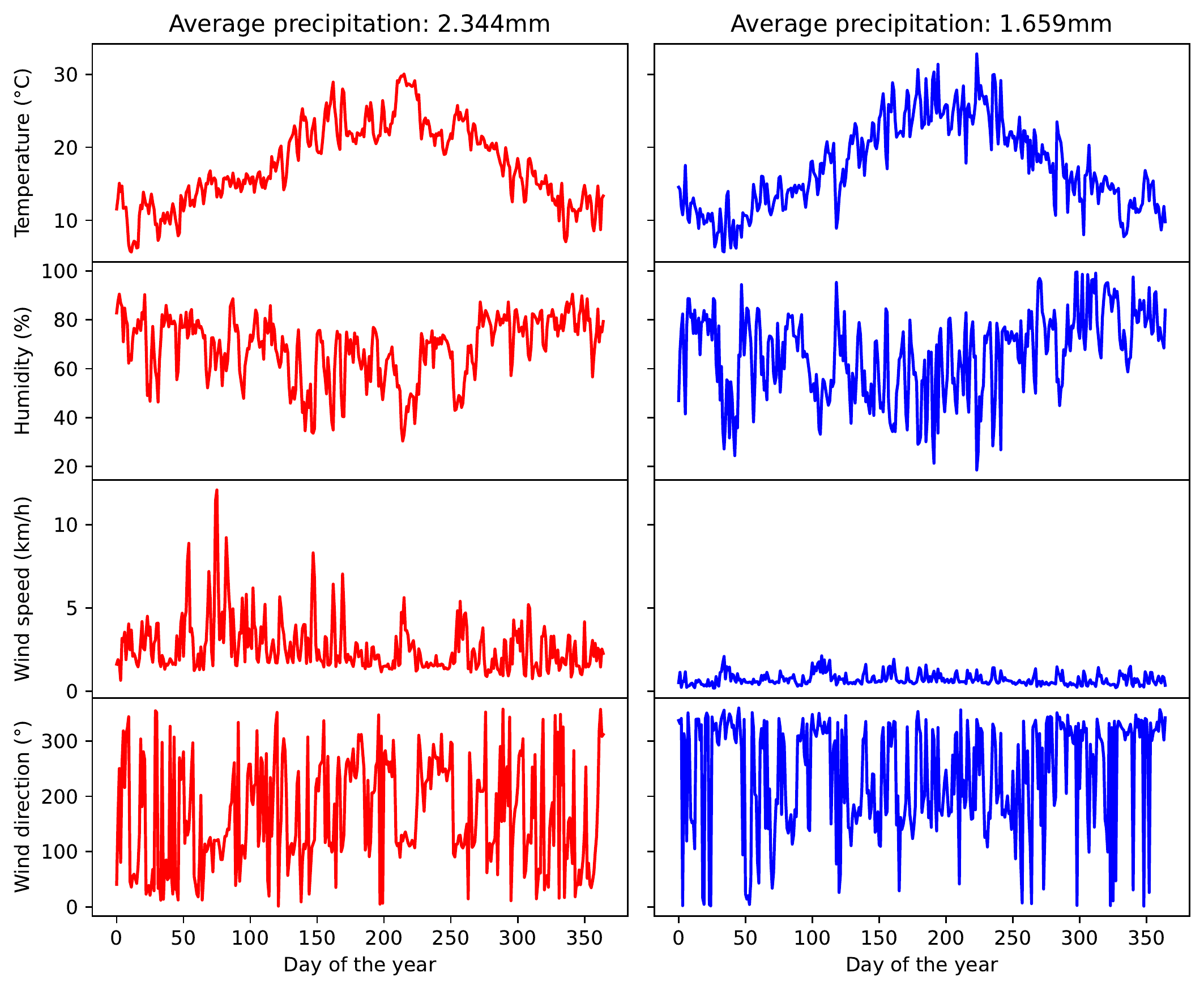}
\caption{Two different years of measurements and the associated average precipitation in the PrecipitationAndalusia dataset}
\label{fig:datasetprecipitation}
\end{figure}

This dataset has been obtained from the Andalusia Government (Spain)\textsuperscript{\ref{PrecipitationAndalusia}}. Data was retrieved from different stations in the 8 districts of Andalusia: Almeria, Cadiz, Cordoba, Granada, Huelva, Jaen, Malaga, and Sevilla, from 2000 until February 2014. This dataset, known as \textbf{PrecipitationAndalusia}, includes the daily averaged temperature, humidity, and wind speed and direction as inputs, whereas the output is the average precipitation. Two examples are illustrated in Figure \ref{fig:datasetprecipitation}.

This resulted in 672 4-dimensional time series of length 365. We randomly sampled 70\% of those time series to use as train data and the remaining 30\% as test data.

\subsubsection{Sierra Nevada mountains measurements}
A dataset published by Osterhuber and Schwartz~\cite{datasets_SIERRANEVADA} and made available on Kaggle\textsuperscript{\ref{SierraNevadaMountainsSnow}} consists of daily measurements of minimum and maximum air temperatures, precipiation, and snowpack characteristics made at a field station in Sierra Nevada, United States, between 1971 and 2019.

We split the measurements into groups of 30 consecutive days and used the minimum and maximum air temperature (in °C) and precipitation (in mm) to create 3-dimensional time series of length 30. Using the amount of new snow (in cm) accumulated during each respective 30-day timeframe as the target variable, we created the \textbf{SierraNevadaMountainsSnow} dataset, with 500 instances in total. Around 20 instances were removed due to missing values. The dataset is split into train and test sets by randomly sampling 30\% of the data as test. This data can be used to train models to predict heavy snowfall and prepare cities and roads for harsh weather conditions.

\hyperfootnotetext{\label{SierraNevadaMountainsSnow}\textsuperscript{\ref*{SierraNevadaMountainsSnow}}\url{https://perma.cc/2KQT-C5AW}}

\subsection{Equipment monitoring}

\subsubsection{Electric motor temperature}

Kirchgässner, Wallscheid and Böcker~\cite{datasets_electricmotor} collected large amounts of sensor data, made available on Kaggle\textsuperscript{\ref{ElectricMotorTemperature}}, from a permanent magnet synchronous motor (PMSM) deployed on a test bench. The dataset consists of multiple measurement sessions, each ranging between one and six hour long, whose recordings were sampled at 2Hz. The sensors collected a variety of features, such as current and voltage, ambient, coolant, and rotor temperatures, and motor speed. Rotor temperature, specifically, is not reliably and economically measurable in a commercial vehicle, thus being an interesting candidate for response variable. The data is, therefore, useful for industrial processes and monitoring.

\begin{figure}[htb]
\centering
\includegraphics[width=\columnwidth]{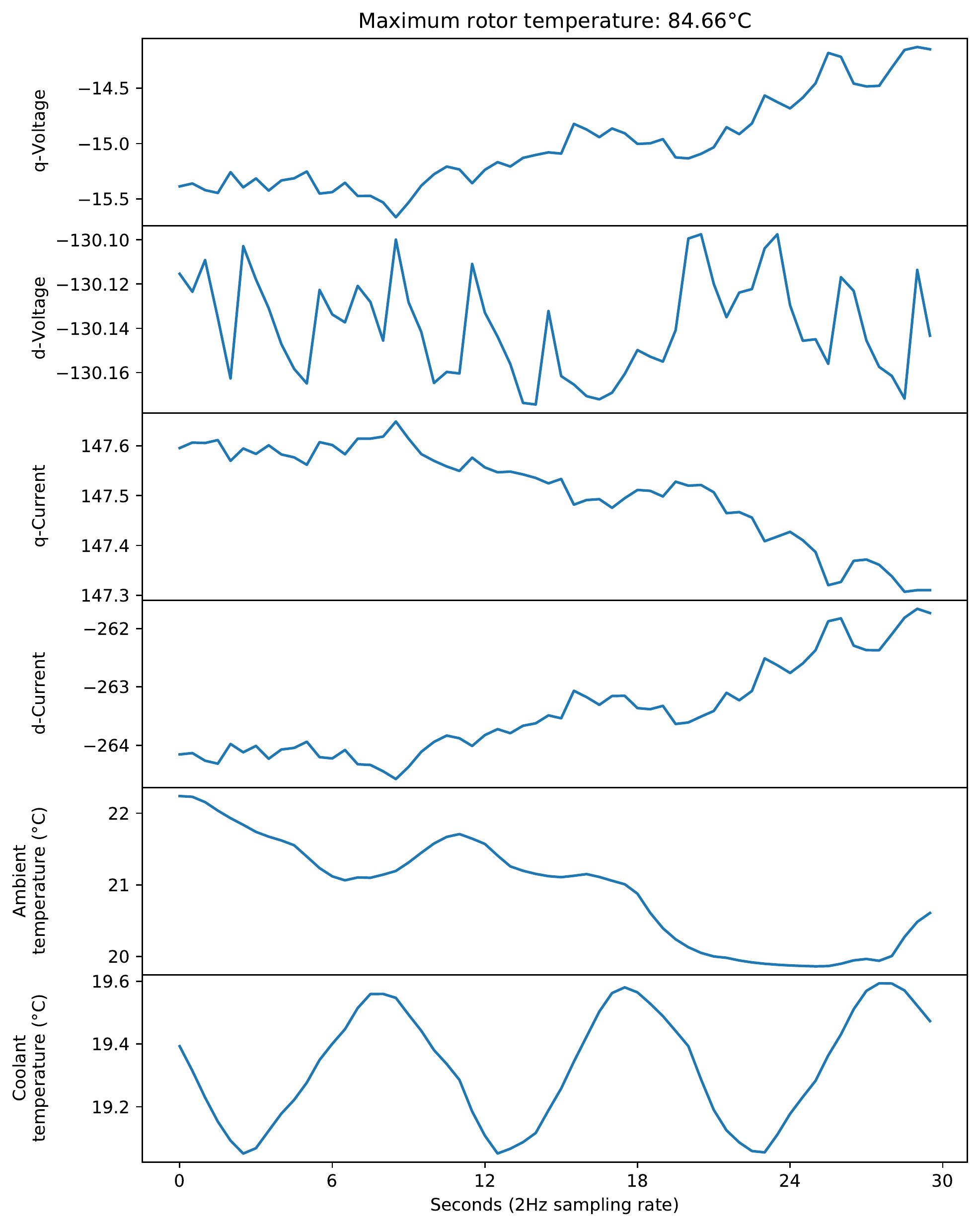}
\caption{Example of currents, voltages, and ambient and coolant temperatures measurements used to predict maximum rotor temperature in the ElectricMotorTemperature dataset}
\label{fig:datasetmotor}
\end{figure}

Therefore, we created the \textbf{ElectricMotorTemperature} dataset by first splitting the measurement sessions into groups of 30 consecutive seconds, i.e. 60 timepoints. Then, we used the recorded ambient and coolant temperatures and d and q components of voltage and current as predictors to form 6-dimensional time series of length 60. The target variable is the maximum recorded rotor temperature during each respective 30-second time frame. An example of such a time series is shown in Figure \ref{fig:datasetmotor}. The resulting dataset has 22148 instances, of which 70\% were sampled as training data and the remaining 30\% as testing data.  

\subsubsection{Home activity monitoring of gases}

\begin{figure}[htb]
\centering
\includegraphics[width=\columnwidth]{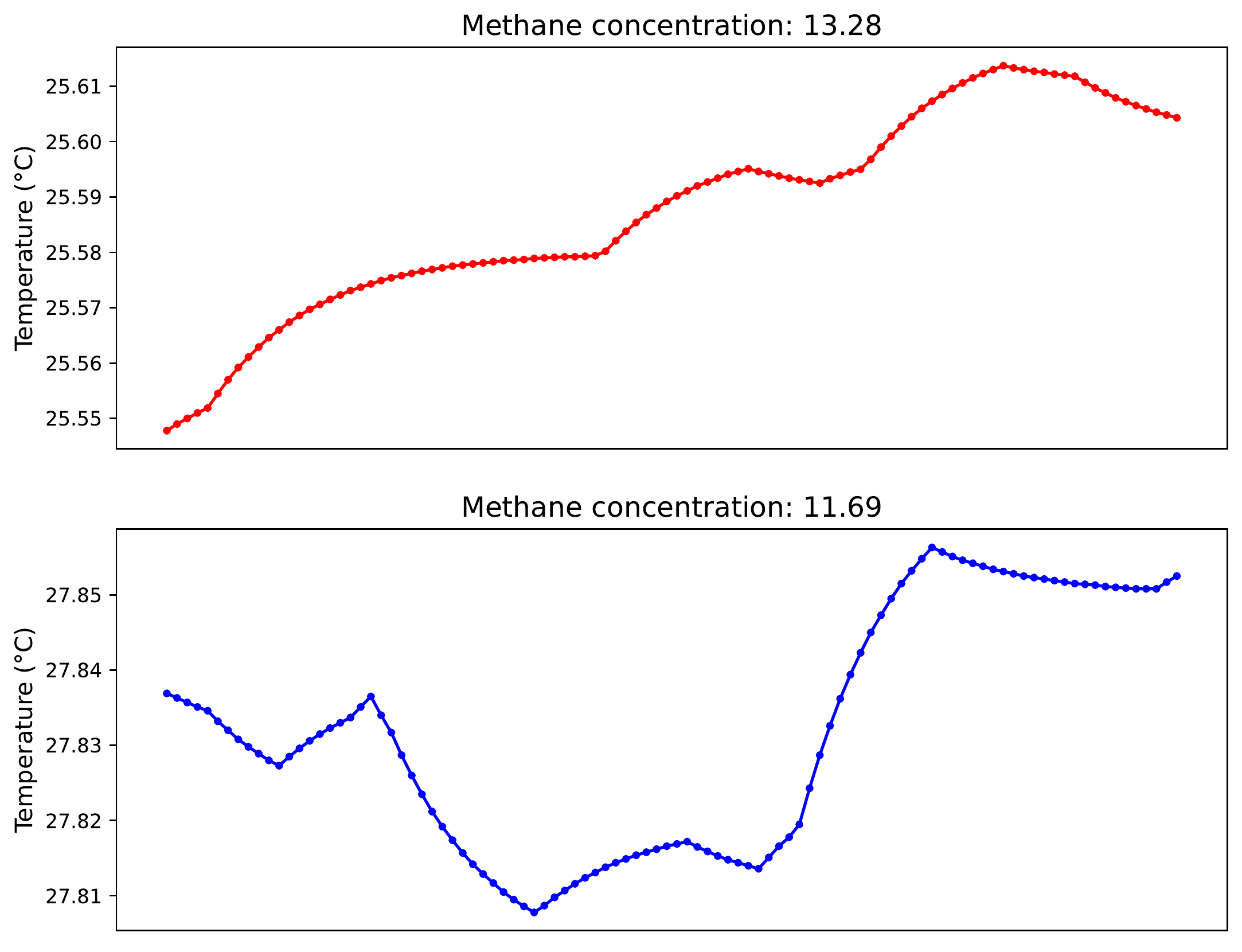}
\caption{Temperature data and respective methane concentration in the MethaneMonitoringHomeActivity dataset}
\label{fig:datasetmethane}
\end{figure}

This dataset has been obtained from UCI Machine Learning Repository\textsuperscript{\ref{GasMonitoringHomeActivity}}. The authors \cite{huerta2016online} collected recordings of a gas sensor array composed of 8 MOX gas sensors, as well as a temperature and humidity sensor. Sensors were exposed to three different conditions: presence of wine, banana and background activity. Two datasets have been collated from this repository: \textbf{LPGasMonitoringHomeActivity} and \textbf{MethaneMonitoringHomeActivity}. The first estimates the liquefied petroleum gas concentration from humidity measurements. On the other hand, the second one estimates the methane concentration from temperature measurements. The latter is illustrated in Figure \ref{fig:datasetmethane}.

This resulted in 2882 univariate time series of length 100. We randomly sampled 70\% of those time series to use as train data and the remaining 30\% as test data.

\hyperfootnotetext{\label{GasMonitoringHomeActivity}\textsuperscript{\ref*{GasMonitoringHomeActivity}}\url{https://perma.cc/69GV-CR7S}}

\subsubsection{Gas sensor array under flow modulation}

Ziyatdinov \textit{et al.}~\cite{datasets_ZIYATDINOV2015538} combined an array of 16 metal-oxide gas sensors and an external mechanical ventilator to simulate sniffling behaviour within the biological respiration cycle. The study extracted high and low frequency features from the signals and proposed a regression problem where the predictors are either of those features and the responses are the concentrations of two analytes used to form test gasses, acetone and ethanol. The data collected by the authors has been made available in the UCI Machine Learning repository\textsuperscript{\ref{GasSensor}}. The development of better regression techniques for this data should lead to improvements in early detection of gases in chemo-sensory systems.

We used the raw signal data with 58 samples, each one consisting of 16 sensors and totalling 928 time series, to create two datasets: \textbf{GasSensorArrayEthanol}, in which the target variable is the ethanol concentration in the tested gas, and \textbf{GasSensorArrayAcetone}, in which the target is similarly the concentration of acetone. Both consist of 464 univariate time series of length 7500 and are split into train and test sets by randomly selecting 30\% of instances as test. 

\hyperfootnotetext{\label{GasSensor}\textsuperscript{\ref*{GasSensor}}\url{https://perma.cc/WD2S-KEV9}}

\subsubsection{Wave elevation and line tension}

A dataset published on Kaggle\textsuperscript{\ref{WaveTensionData}} consists of a simulation of a ship. The dataset's goal is to predict the tension of a string given temporal wave elevation data. While the data is simulated, well-fitted models should improve monitoring of naval equipment during harsh conditions.

Thus, we created the \textbf{WaveTensionData} dataset by separating the source data into univariate time series of length 57 and using wave height as the predictor and the corresponding average string tension as the target variable. The resulting dataset has 1893 instances, of which 70\% were sampled as training data and the remaining 30\% as testing data. 

\subsection{Health monitoring}

\subsubsection{Bar Crawl: Detecting Heavy Drinking Episodes}

\begin{figure}[htb]
\centering
\includegraphics[width=\columnwidth]{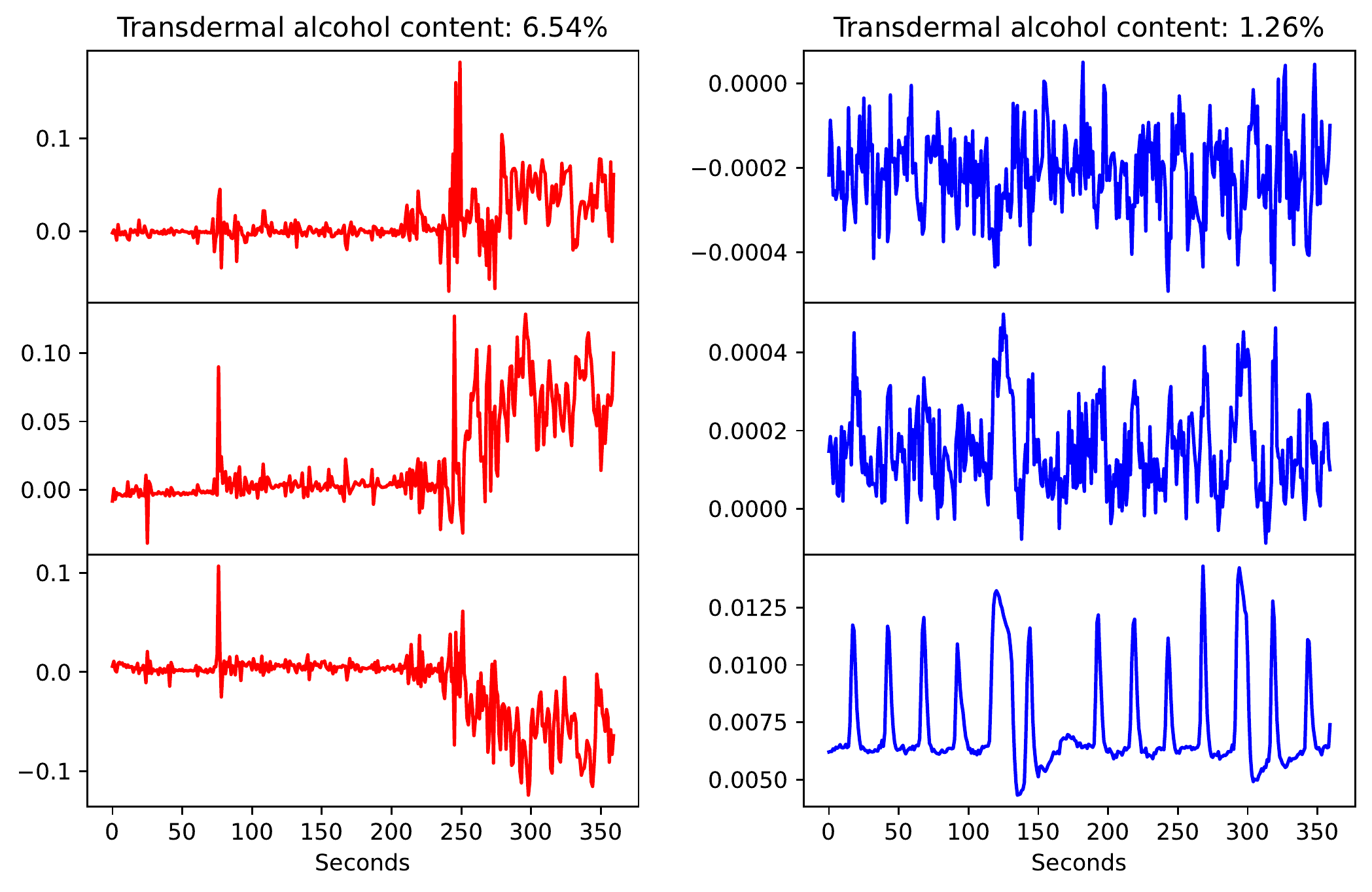}
\caption{Two samples of accelerometer data and alcohol concentration in the BarCrawl6min dataset}
\label{fig:datasetbarcrawl}
\end{figure}

This dataset, made available in the UCI Machine Learning repository\textsuperscript{\ref{BarCrawl6min}} by their authors \cite{killian2019learning}, consists in predicting heavy drinking episodes via mobile data. Data was collected from smartphones from 13 participants. The goal is to estimate the transdermal alcohol content by using an accelerometer. We have, therefore, created the \textbf{BarCrawl6min} dataset, where each input dimension corresponds to a different axis of the accelerometer. Moreover, even though data was recorded at 30 minutes intervals, in order to accurately estimate the drinking episode, only the last 6 minutes from the recording were kept. Two resulting samples are shown in Figure \ref{fig:datasetbarcrawl}. Note that all data is fully anonymised and that TAC readings were preprocessed/cleaned by the authors.

This resulted in 201 3-dimensional time series of length 360, which correspond to 6 minutes of secondly measurements. We randomly sampled 70\% of those time series to use as train data and the remaining 30\% as test data.

\hyperfootnotetext{\label{BarCrawl6min}\textsuperscript{\ref*{BarCrawl6min}}\url{https://perma.cc/7VP5-79LN}}

\subsubsection{Covid-19 in Andalusia}

\begin{figure}[htb]
\centering
\includegraphics[width=\columnwidth]{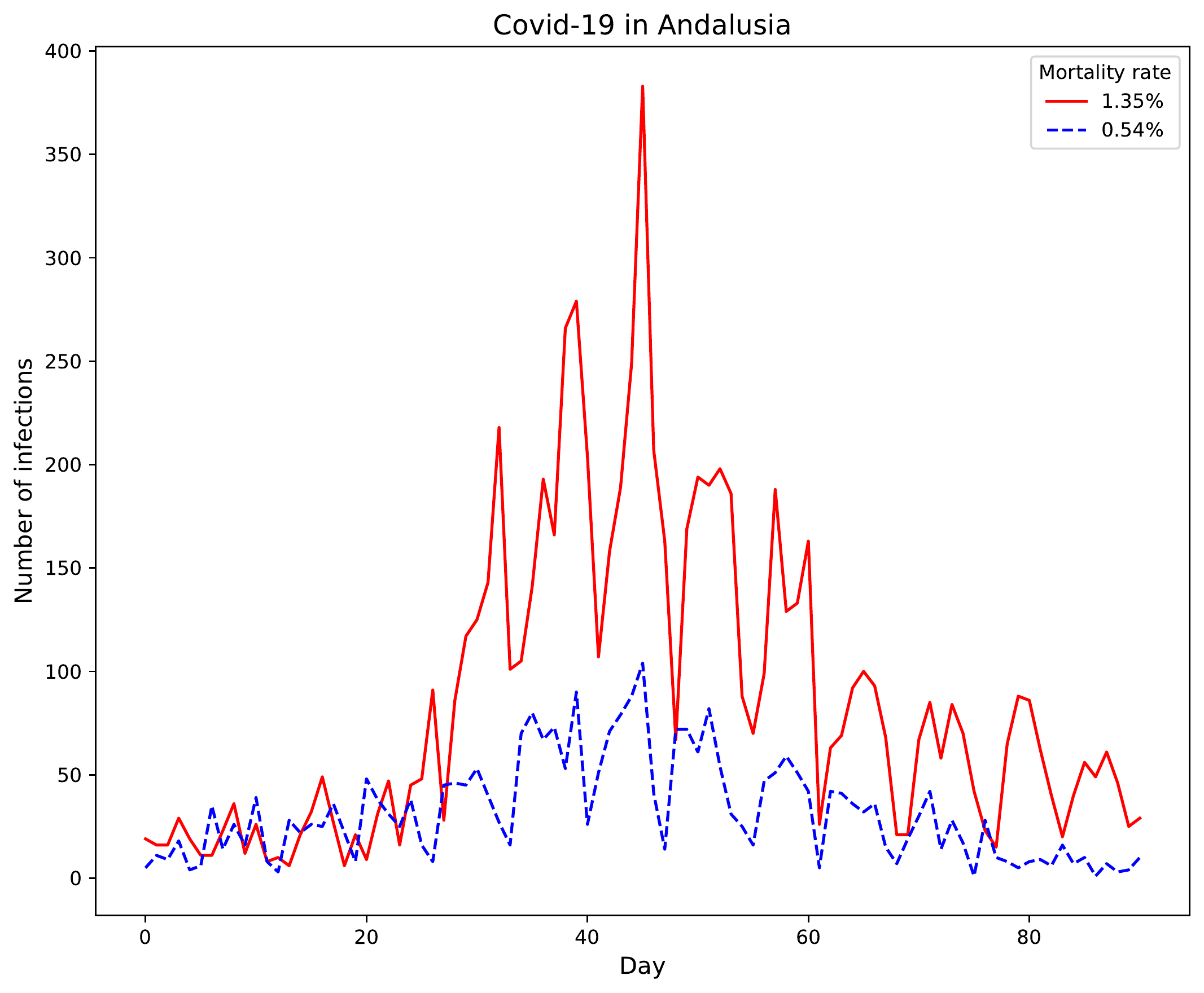}
\caption{Two different Covid-19 waves and their mortality rates in the Covid19Andalusia dataset}
\label{fig:datasetcovidandalusia}
\end{figure}

This dataset consists of estimating the mortality rate during Covid-19 waves and in different districts in eight different areas of Andalusia, its country's second largest and most populated autonomous region, located in southern Spain. The output is the number of deaths in proportion to the total number of infected people in that district. This dataset, known as \textbf{Covid19Andalusia}, has been made public by the authors of the work \cite{diaz2022covid}, who took the data available in the Andalusia Government Website\textsuperscript{\ref{Covid19Andalusia}}. This dataset comprises 6 waves from a total of 34 districts. All waves are equal-length (91 points) since we have considered 45 days before and after the peak of the outbreak, as it has been demonstrated to be the most relevant data. Two such waves are illustrated in Figure \ref{fig:datasetcovidandalusia}.

The aforementioned dataset includes 204 unidimensional time series of length 91. 70\% of those 204 time series have been randomly selected for the training set, whereas the remaining 30\% belong to the testing set.

\hyperfootnotetext{\label{Covid19Andalusia}\textsuperscript{\ref*{Covid19Andalusia}}\url{https://perma.cc/U55Q-U4KZ}}

\subsubsection{Pressure of a ventilator connected to a sedated patient's lung}

This dataset was made available by Kaggle in collaboration with Google Brain\textsuperscript{\ref{VentilatorPressure}}. Data used in this competition was produced by connecting a ventilator to an artificial bellows test lung through a respiratory circuit. The goal is to estimate the pressure, a value ranging from 0 to 100, representing how much the inspiratory solenoid valve is open to let air into the lung. The proposed dataset, named \textbf{VentilatorPressure}, includes time series corresponding to approximately 3-second breaths. Dimensions are the control input and output for the inspiratory solenoid valve. 

The final dataset includes 75450 time series with 2 dimensions and length 80. The training dataset comprises randomly sampled 70\% of the series, whereas the remaining 30\% belong to the testing dataset.

\subsection{Sentiment analysis}

\subsubsection{Cryptocurrency sentiment}

By combining historical sentiment data for 4 cryptocurrencies, extracted from EODHistoricalData\textsuperscript{\ref{CryptoSentiment2}} and made available on Kaggle\textsuperscript{\ref{CryptoSentiment}}, with historical price data for the same cryptocurrencies, extracted from CryptoDataDownload\textsuperscript{\ref{CryptoSentiment3}}, we created the \textbf{BitcoinSentiment}, \textbf{EthereumSentiment}, \textbf{CardanoSentiment}, and \textbf{BinanceCoinSentiment} datasets, with 332, 356, 107, and 263 total instances, respectively.

In all four datasets, the predictors are hourly close price (in USD) and traded volume for each respective cryptocurrency during a day, resulting in 2-dimensional time series of length 24. The response variable is the normalized sentiment score on the day spanned by the timepoints. The datasets were split into train and test sets by randomly selecting 30\% of each set as test data. Using this data, companies can better prepare for shifts of public perception regarding cryptocurrencies.

\hyperfootnotetext{\label{CryptoSentiment2}\textsuperscript{\ref*{CryptoSentiment2}}\url{https://perma.cc/37GN-BMRL}}

\hyperfootnotetext{\label{CryptoSentiment3}\textsuperscript{\ref*{CryptoSentiment3}}\url{https://perma.cc/4M79-7QY4}}

\subsubsection{Sentiment on natural gas prices}

Natural gas prices historical data was taken from the U.S. Energy Information Administration\textsuperscript{\ref{NaturalGasPriceSentiment}} along with corresponding sentiment scores obtained by analysing relevant tweets on the topic. From this data, we created the \textbf{NaturalGasPricesSentiment} dataset.

\begin{figure}[htb]
\centering
\includegraphics[width=\columnwidth]{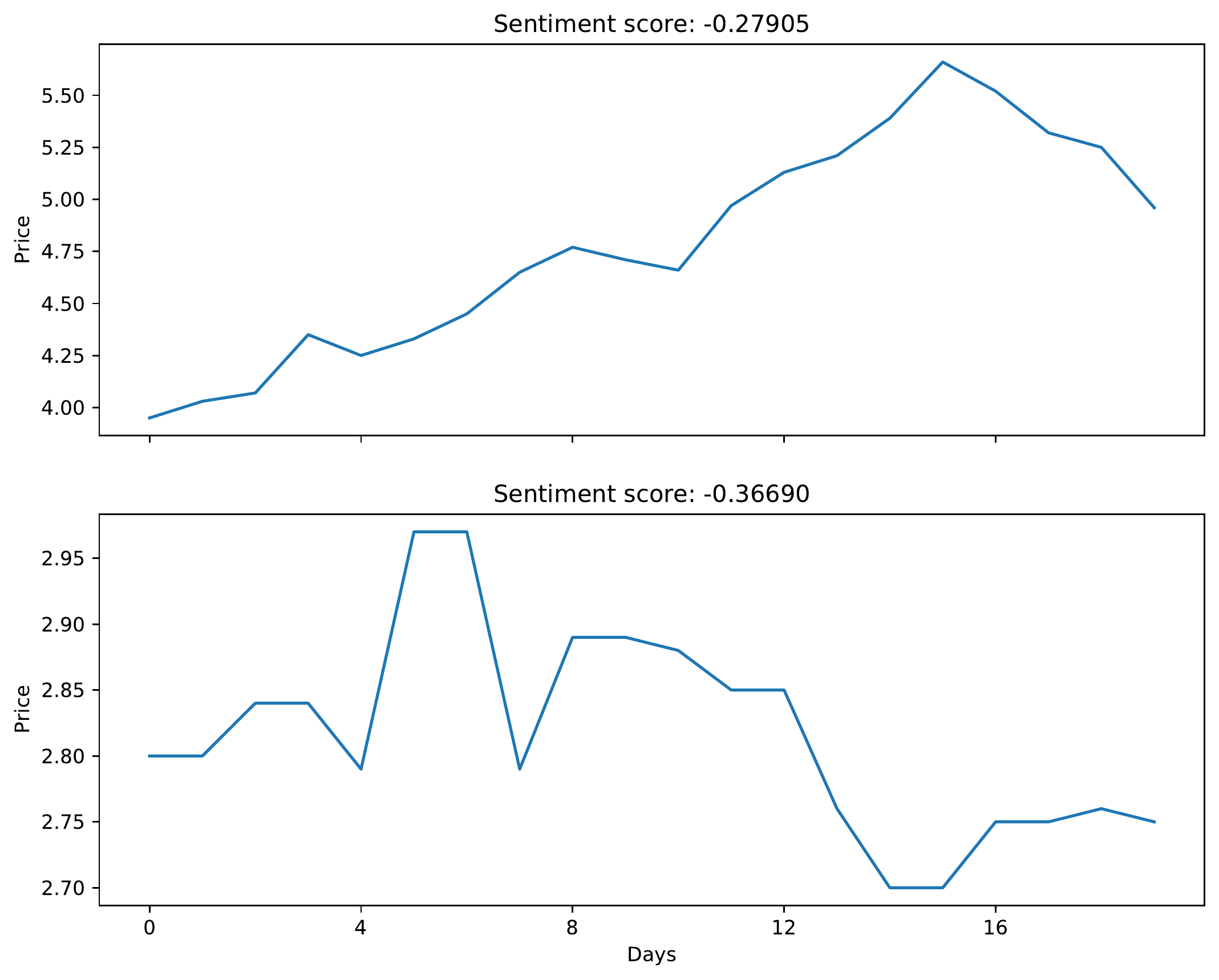}
\caption{Examples of gas price time series from the NaturalGasPriceSentiment dataset and the respective sentiment score}
\label{fig:datasetnaturalgas}
\end{figure}

We first split the data into groups of 20 consecutive business days. We then used the daily natural gas prices as predictors and the average sentiment score during each 20-day time frame as the response variable. The final dataset has 93 univariate time series of length 20, of which 70\% were randomly sampled as training data and the remaining 30\% as testing data. Two of those time series are shown in Figure \ref{fig:datasetnaturalgas}. Again, companies and local governments can use the data to analyse and predict shifts in public perception on natural gas.

\hyperfootnotetext{\label{NaturalGasPriceSentiment}\textsuperscript{\ref*{NaturalGasPriceSentiment}}\url{https://perma.cc/8AP5-5R7R}}

\section{Regressor configurations}

Parameter settings for all algorithms is shown in Table \ref{tab:parameters}. 

\begin{table}[h]
    \caption{Regressor configurations for our experiments where $m$ is the series length, $d$ is the
number of dimensions and $rm$ is the lengths of DrCIF representations.}
    \label{tab:parameters}
    \resizebox{0.45\textwidth}{!}{
    \begin{tabular}{c|l}
        \toprule \toprule 
        Regressor & Configuration \\
        \midrule
        Ridge & Regularization strength: $ \in \{1\cdot10^{-4}, 1\cdot10^{-3}, \ldots, 1\cdot10^{4}\}$ \\ [0.3cm]
        \multirow{3}{*}{grid-SVR} & Kernel $ \in  \{\text{RBF}, \text{Sigmoid}\}$ \\
        & Regulatisation parameter $ \in \{1\cdot10^{-2}, 1\cdot10^{-1}, \ldots, 1\cdot10^{2}\}$ \\
        & Kernel Coefficient $ \in \{1\cdot10^{-4}, 1\cdot10^{-1}, \ldots, 1\cdot10^{-1}\}$ \\ [0.3cm]
        RandF & Num. Estimators: $500$ \\ [0.3cm]
        RotF & Num. Estimators: $500$ \\ [0.3cm]
        \multirow{2}{*}{XGBoost} & Num. Estimators: $500$ \\
        & Learning Rate: $0.1$ \\ [0.3cm]
        FPCR & FPCA Num. Components: $10$ \\ [0.3cm]
        \multirow{4}{*}{FPCR-Bs} & Smooth: B-splines \\
        & FPCA Num. Components: $10$ \\
        & Num. Basis Functions: $10$ \\
        & Order: $4$ \\ [0.3cm]
        ROCKET & Num. Kernels: $10000$ \\ [0.3cm]
        \multirow{3}{*}{MultiROCKET} & Num. Kernels: $6250$ \\
        & Max. Num. Dilations per Kernel: $32$ \\
        & Num. Features per Kernel: $4$ \\ [0.3cm]
        $1$NN-ED & Num. Neighbours: $1$ | $5$ \\
        $5$NN-ED & Distance measure: Euclidean \\ [0.3cm]
        $1$NN-DTW & Num. Neighbours: $1$ | $5$ \\
        $5$NN-DTW & Distance measure: DTW \\ [0.3cm]
        FreshPRINCE & Num. Estimators: $500$ \\ [0.3cm]
        \multirow{2}{*}{TSF} &  Num. Estimators: $500$ \\ 
        & Num. Intervals per Tree: $\sqrt{m}$ \\ [0.3cm]
        \multirow{3}{*}{DrCIF} & Num. Estimators: $500$ \\  
        & Num. Intervals per Representation: $4+(\sqrt{d}\sqrt{rm})/3$ \\
        & Num. Features per Tree: $10$ \\[0.3cm]
        \multirow{4}{*}{CNN} & Num. Epochs: $2000$ \\
        & Batch size: $16$ \\
        & Kernel size: $7$ \\
        & Num. Convolutional Layers: $2$ \\ [0.3cm]
        \multirow{2}{*}{FCN} & Num. Epochs: $2000$ \\
        & Batch size: $16$ \\ [0.3cm]
         & Num. Components: $1$ (inceptionE: $5$) \\
        Inception & Num. Epochs: $1500$ \\
        InceptionE & Batch size: $64$ \\
        & Kernel size: $40$ \\ [0.3cm]
        \multirow{2}{*}{ResNet} & Num. Epochs: $1500$ \\
        & Batch size: $16$ \\
        \bottomrule \bottomrule
    \end{tabular}
    }
\end{table}

\section{Results}
RMSE results for the best performing regressors on 63 TSER datasets are available in the accompaning website\footnote{\url{https://tsml-eval.readthedocs.io/en/latest/publications/2023/tser_archive_expansion/tser_archive_expansion.html}.}.

\bibliographystyle{IEEEtran}
\bibliography{IEEEabrv,bibliography}

\vspace{-33pt}
\begin{IEEEbiography}[{\includegraphics[width=1in,clip,keepaspectratio]{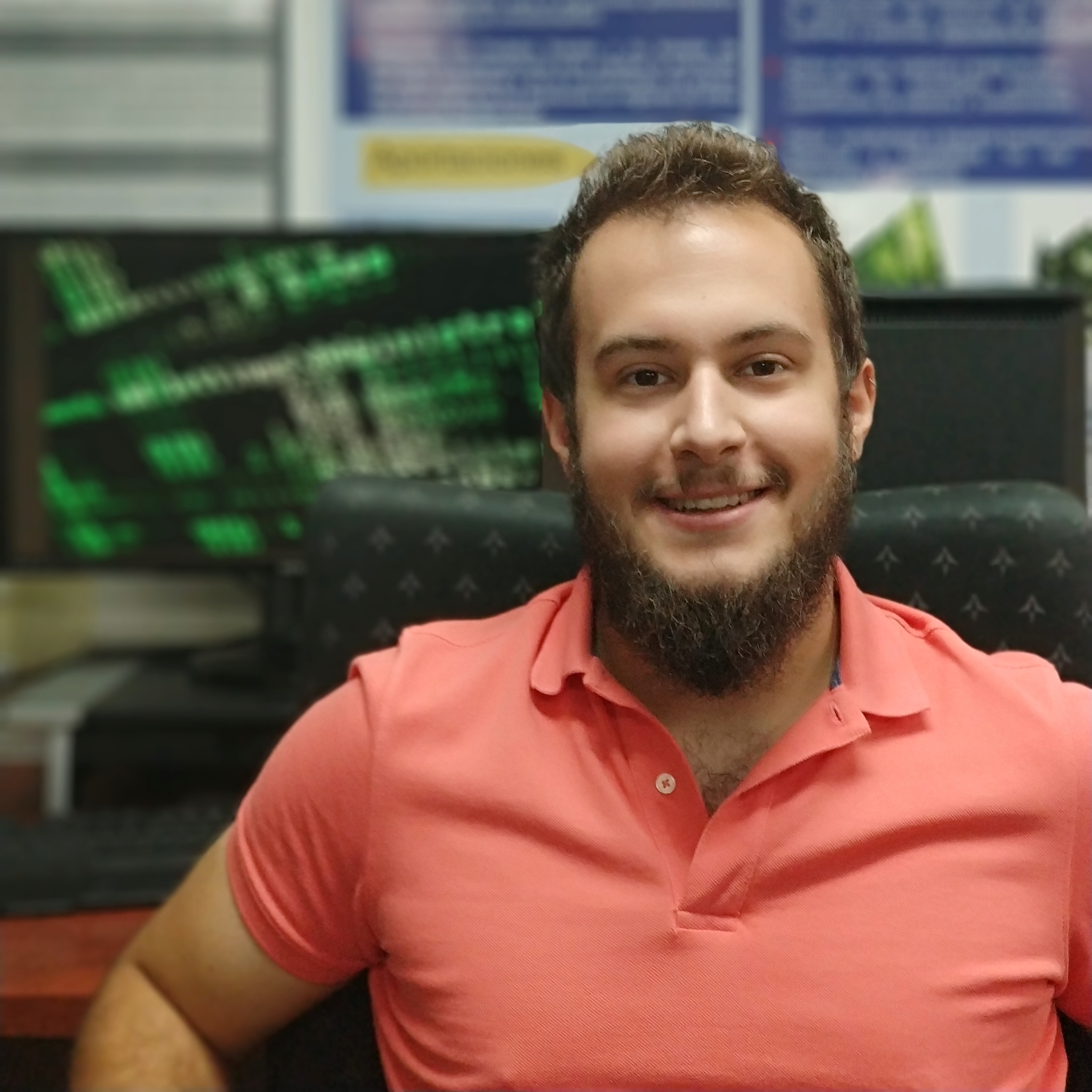}}]{David Guijo-Rubio}
received the BSc degree in Computer Science in 2016 and the MSc degree in Artificial Intelligence in 2017. In 2021, he received the PhD degree in Computer Science from the University of Córdoba (Spain). He is working as an Postdoctoral Researcher at both the University of Córdoba and the University of East Anglia (UK). His current interests include ordinal classification, pattern recognition and different tasks applied to time series (classification -nominal and ordinal, clustering and extrinsic regression). \end{IEEEbiography}
\vspace{-33pt}
\begin{IEEEbiography}[{\includegraphics[width=1in,height=1.25in,clip,keepaspectratio]{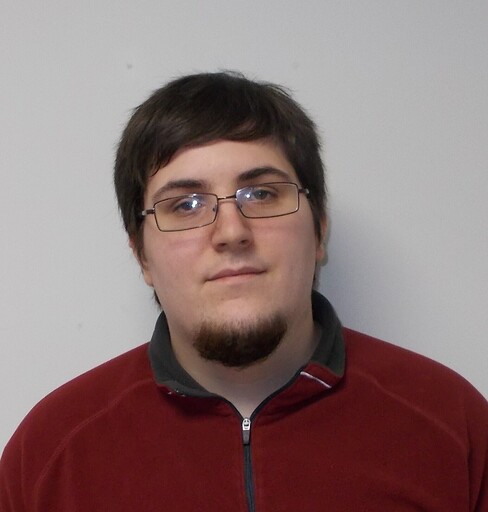}}]{Matthew Middlehurst}
received the BSc degree in Computer Science in 2018 and the PhD degree in Computer Science in 2023 from the University of East Anglia (UEA), Norwich, UK. He is working as a Senior Research Associate at UEA in the time series machine learning group. His research interests are primarily time series classification related, with recent expansion to time series clustering and time series extrinsic regression. He is interested in the provision and maintenance of open source software for researchers in time series based fields.\end{IEEEbiography}
\vspace{-33pt}
\begin{IEEEbiography}[{\includegraphics[width=1in,height=1in,clip,keepaspectratio]{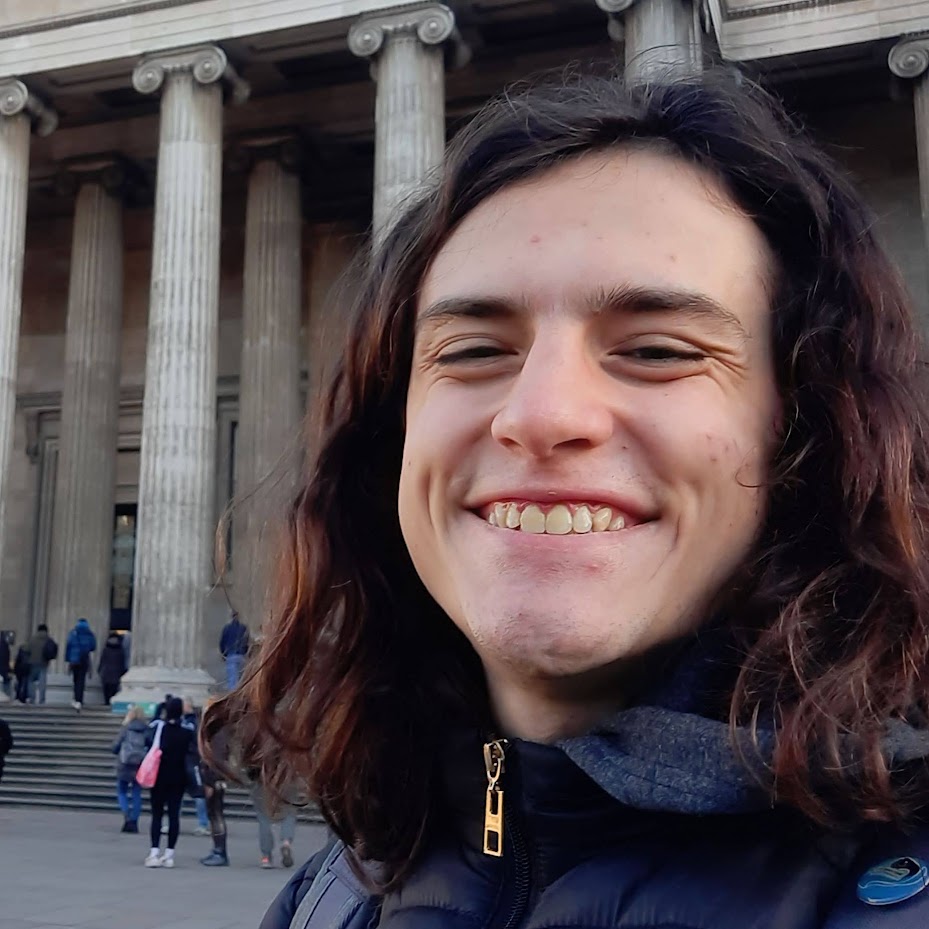}}]{Guilherme Arcencio}
is pursuing the BSc in Computer Engineering from the Federal University of São Carlos (UFSCar), São Carlos, Brazil. He worked as a Research Intern at the University of East Anglia (UEA) in the time series machine learning group. His current research interests include time series machine learning with a focus on extrinsic regression, as well as the development and maintenance of open source software. \end{IEEEbiography}
\vspace{-33pt}
\begin{IEEEbiography}[{\includegraphics[width=1in,height=1in,clip,keepaspectratio]{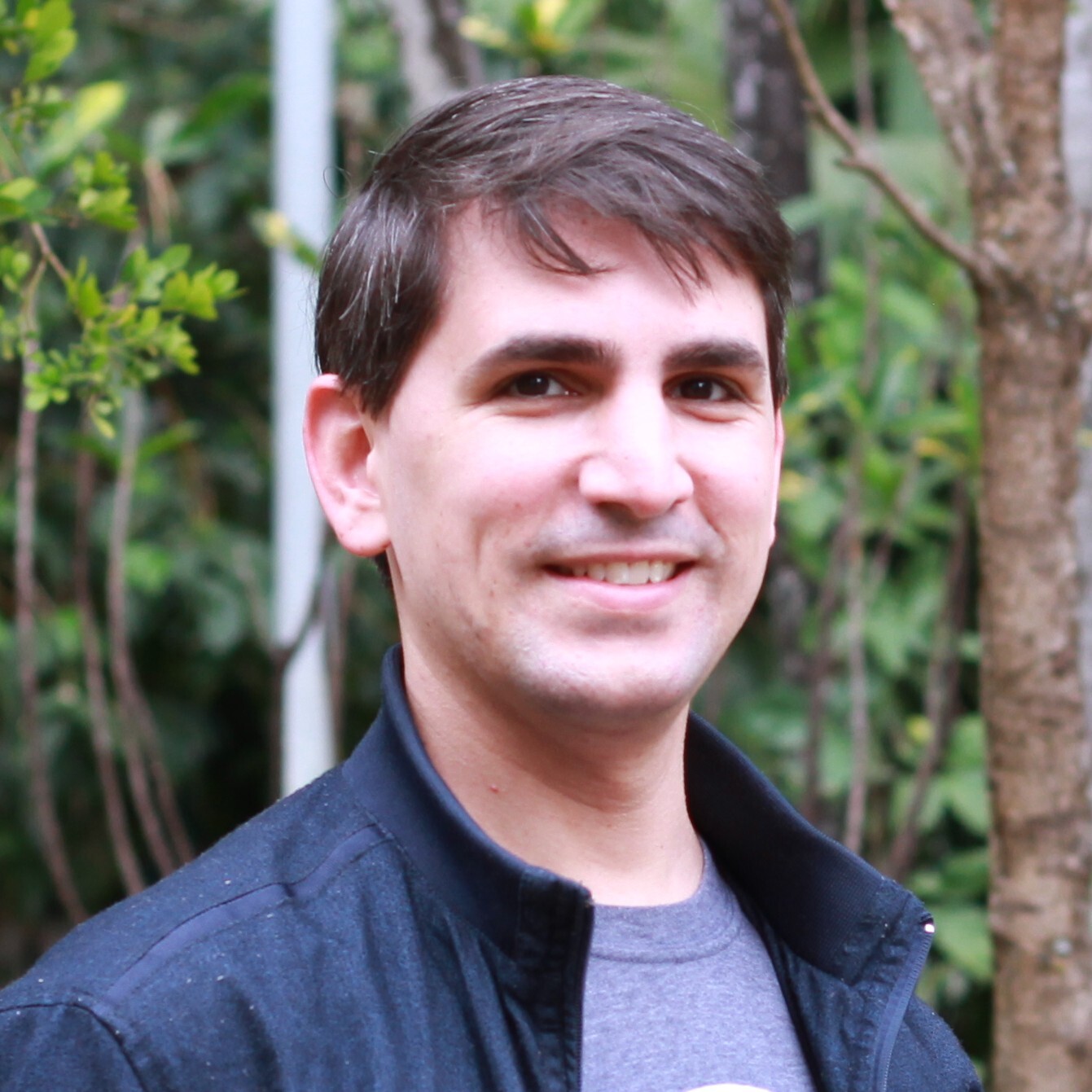}}]{Diego Furtado Silva}
received the BSc degree in Computer Science at the Institute of Mathematics and Computer Sciences (ICMC), University of São Paulo (USP), São Carlos, as well as the MSc and PhD degree in Computer Science and Computational Mathematics, in time series and digital signal analysis. Currently, he is an Assistant Professor at ICMC-USP and collaborate with the Computational Intelligence Laboratory (LABIC) and the Data Mining and Application Group at the Federal University of São Carlos (MIDAS-UFSCar). His main research interests are machine learning foundations and applications. \end{IEEEbiography}
\vspace{-33pt}
\begin{IEEEbiography}
[{\includegraphics[width=1in,height=2in,clip,keepaspectratio]{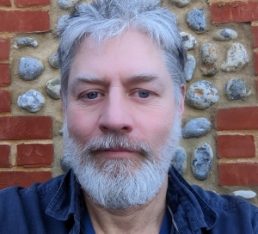}}]{Anthony Bagnall}
received the PhD degree in Computer Science from the University of East Anglia (UEA), Norwich, UK in 2001. Since 2018 he has been a Full Professor of Computer Science at UEA, where he leads the time series machine learning group. His primary research interest is in time series machine learning, with a historic focus on classification, but more recently looking at clustering and regression. He has a side interest in ensemble design. \end{IEEEbiography}

\vfill

\end{document}